\documentclass[12pt]{article}
\usepackage{wrapfig}
\usepackage[noend]{algpseudocode}
\usepackage{algorithm}

\usepackage[english]{babel}

\usepackage[letterpaper,top=2.5cm,bottom=2.5cm,left=2.5cm,right=2.5cm]{geometry}

\usepackage{amsfonts}
\usepackage{amsmath}
\newcommand{\sign}{\text{sign}}
\usepackage{graphicx}
\usepackage{kbordermatrix,etoolbox}
\usepackage{blkarray}
\usepackage[colorlinks=true, allcolors=blue]{hyperref}
\usepackage{natbib}
\usepackage{multirow}
\usepackage{graphicx}
\DeclareMathAlphabet{\mathcal}{OMS}{cmsy}{m}{n}
\usepackage{xcolor}

\newcommand{\added}[1]{\textcolor{black}{#1}}
\usepackage{xcolor}
\usepackage{soul}
\usepackage{eurosym}
\usepackage{authblk}
\usepackage{tikz}
\newcommand{\killpunct}[1]{}

\usetikzlibrary{arrows.meta,arrows}

\usepackage{titlesec}
\titleformat*{\subsubsection}{\normalfont\itshape}

\usepackage{tikzscale}
\title{A Cybersecurity Risk Analysis Framework for Systems with Artificial
Intelligence Components}
\author[1]{J.M. Camacho}
\author[2]{A. Couce-Vieira}
\author[3]{D. Arroyo}
\author[1]{D. Rios Insua}
\affil[1]{Inst. Mathematical Sciences, ICMAT-CSIC, Spain}
\affil[2]{University of Vigo, Spain}
\affil[3]{Inst. Information Technology, ITEFI-CSIC, Spain}

\date{  }
\usepackage{setspace}
\begin{document}

\maketitle

\begin{abstract}
The introduction of the European Union Artificial Intelligence Act, the NIST Artificial Intelligence Risk Management Framework,  
and related norms demands a better understanding and implementation 
of novel risk analysis 
approaches to evaluate systems with Artificial Intelligence components. This paper provides a 
  cybersecurity risk analysis framework that can help assessing such systems. We use an 
  illustrative  
   example concerning automated driving systems.
\end{abstract}


\pagebreak
\section{INTRODUCTION}

Artificial Intelligence (AI) is broadly understood as \textcolor{black}{the study and implementation of computer systems capable of carrying out activities that would typically need human intelligence}. Its importance 
 is immediately appreciated if we think of its  
application to speed up processes that previously took years of expensive research, 
as in drug discovery e.g.\
\cite{gallego2021ai}; or to facilitate the introduction of radically new technologies like automated driving systems \textcolor{black}{(ADSs)}, e.g.\ \cite{caballero2021decision}. Huge investments by the European Union (EU), USA or China, and big tech firms in AI also showcase its potential. Yet alongside its benefits, the introduction of AI entails risks. Examples include deep fake technology 
  employed to insert celebrities' faces onto pornographic content \citep{hasan2019combating}, the use of AI to facilitate the generation of biochemical weapons \citep{urbina2022dual}, or the creation of fake content preventing cyber-attribution \citep{leone2023spiral}. Moreover, risks are exacerbated by the growing speed of AI progress.  A popular report by  the 
Center for Research on Foundation Models (CRFM) \citep{Stanford2021FoundationModels}
emphasises the risks associated with such type of \textit{ models}, large-scale, pre-trained deep learning networks providing multi-purpose AI, usually based on unprecedented natural language processing (NLP) capabilities. 

To shed some light on the production and deployment of AI-based systems, in 
particular their safety and security, the EU has approved the {\em AI Act}  \citep{commissie2021proposal} (from now on the Act), the first law 
 globally regulating AI. However, implementing the Act posits several challenges  made evident by, e.g., ChatGPT \citep{helberger2023chatgpt}. The Act classifies AI systems into four risk categories (unacceptable, high, limited or minimal) accordingly demanding appropriate remedial actions \citep{madiega2023artificial}. 
The U.S. National Institute of Standards and Technology (NIST) has also released an initial version of an AI Risk Management Framework \citep{nist2022ai} (AIRMF from now on) aimed to  \textcolor{black}{support how to manage risks in the design, development, usage, and assessment of AI systems and services.} \textcolor{black}{The {\em Executive Order on the Safe, Secure, and Trustworthy Development and Use of Artificial Intelligence} \citep{whitehouse} has also been recently published by the US government.} 
From a policy perspective, it is evident that AI risk management is attracting lots of attention \citep{agarwala2020supervisory}. 
However, most current approaches remain at a qualitative level, frequently adopting  at most risk matrices 
\citep{ENISA2023} for risk analysis purposes, despite well-known shortcomings \citep{anthony2008s}. Clearly it is of capital importance to conduct solid and trustworthy risk analysis to avoid the manipulation of misinformation attempts that could be weaponized as part of disinformation campaigns \citep{THEKDI2023106129}. 

Given the importance of the issue, we provide here a framework for cybersecurity risk analysis  in systems containing AI components. First, Section 2 remarks the novel risk analysis issues that AI 
systems and components bring in. Section 3 then describes how such issues may be addressed, 
 illustrated with an example concerning  ADS cybersecurity in Section 4. We conclude by discussing some implications of our proposal. Software to reproduce the case study results may be found in \url{https://github.com/***}.\footnote{\tt Blinded for review process.}

\section{NOVEL RISK ANALYSIS ISSUES IN  \\ SYSTEMS WITH AI COMPONENTS}

\begin{figure}[H]
\includegraphics[width=\textwidth]{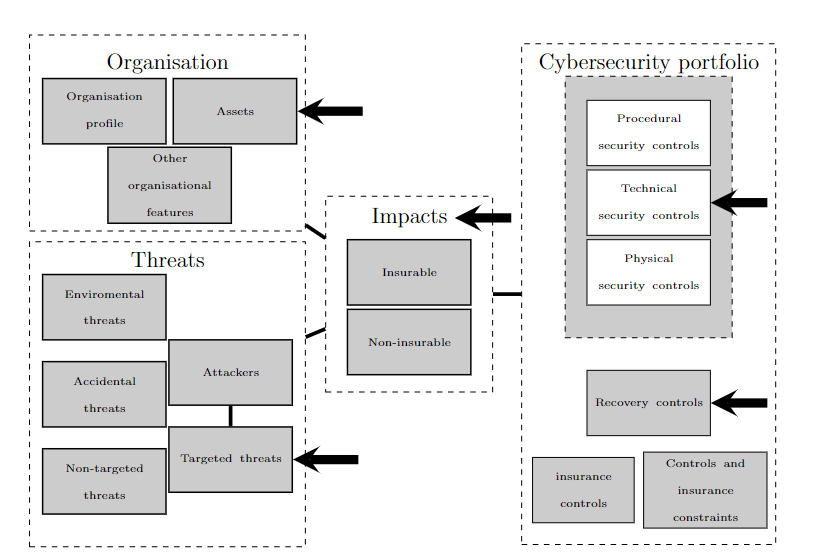}
\caption{Main components in a cybersecurity risk management framework. 
Black arrows indicate AI-affected elements. Adapted from \cite{rios2021adversarial}.}\label{kk}
\end{figure}

Our approach stems from  
the cybersecurity risk analysis framework elaborated in \citeauthor{rios2021adversarial} (\citeyear{insua2020security,rios2021adversarial}), based on the \cite{isf2003}
proposal depicted in Figure \ref{kk}. The framework covers four basic  elements:  \textit{organization features}, relevant \textit{threats} and \textit{impacts}, and the adopted \textit{cybersecurity portfolio}. The figure provides some details concerning such components, e.g.\ impacts are segregated as insurable or non-insurable. Black arrows identify novel risk issues  brought in by the introduction of AI components as we specify after a brief overview.

\paragraph{Overview.}

AI systems, importantly those described as foundation models, bring along new risks and increase others (CRFM, \citeyear{Stanford2021FoundationModels}). 
Their intricacy complicates understanding and control, bringing additional issues in terms of explainability, reliability, safety and accountability, among others.

Emergent \added{capabilities} of AI entail challenges. In some cases, such \added{capabilities} are discovered only after the system has been deployed, notably
 in \added{Large Generative AI models (LGAIMs)}, like ChatGPT, Stable Difussion or Bard, which have demonstrated impressive results on a wide variety of NLP, reasoning and creative tasks. Dozens of emergent abilities have been described, including multi-step arithmetic, or operating other AIs to generate code. Yet they may also \textit{hallucinate}, producing seemingly plausible but untrue statements. The emergence of abilities is also affected by the fact that AIs are goal-oriented systems in the sense that the goals specified by AI developers (upstream) and users (downstream) affect how the 
  system learns, its functionalities and interactions. Moreover, additional risks 
  arise from {\em function creep}, enabling non-valid uses affecting the dependability of the system, and even unsafe, unfair or malicious actions.  
  
  Although risk analysis typically pays attention to emergent, or potentially emergent, features of systems (e.g., failures, vulnerabilities, threats), the existence of emergent \added{capabilities} in AI adds a layer of complexity. In addition, the Act emphasises the so-called {\em high-risk AI systems}, including those used as {\em safety components} of a product or for purposes such as biometric identification, categorization of persons, and critical infrastructure protection.

\paragraph{AI-related impacts.}

Beyond conventional CIA (Confidentiality, Integrity, and Availability) technical impacts \citep{ham2021toward},  \cite{couce2020assessing} (CV20 from now on) present a broader vision of cybersecurity objectives, including business and societal elements. However, several organizations have proposed additional impacts
    brought in by AI relevant in cybersecurity terms. To wit, let us mention the AIRMF and its technical (accuracy, reliability, robustness, resilience),
socio-technical (explainability, interpretability, privacy, safety, bias manageability)
and guiding (fairness, accountability, transparency) principles
from which novel impacts emerge.
 Notably, questions about safety, ethics, bias, privacy and fairness have become a top concern in the digital world,  including AI as it becomes a more popular technology,  and there is a need to adjust, match and, possibly, update earlier 
cybersecurity objectives to cover the new impacts. 

\paragraph{AI-based assets.} 
Many systems include AI blocks, or functional components heavily dependent on AI, that are susceptible to attacks, in particular, the above-mentioned safety components in the Act. A prime contemporary example are ADSs which
have as core components perception systems typically based on classifiers using convolutional neural networks \citep{gallego2022current}. As \cite{boloor2019simple} describes, these have been the subject of, e.g., data poisoning attacks with potentially catastrophic outcomes.
  Following \cite{THEKDI2023106129} and \cite{ETSI}, and given the relevance 
of hardware weaknesses, we adopt a broad view and consider as well hardware 
assets backing AI components.

\paragraph{AI-based security and recovery controls.}
Many systems include AI elements as part of their 
security controls or safety components, susceptible to being fooled with malicious purposes. One example are content filters, aimed to filtrate undesirable content. 
For instance, spam detectors have been shown to be easily fooled with carefully crafted emails \citep{naveiro2019adversarial} and malware detectors \citep{redondo2020protecting} are plagued with 
obfuscated malware. AI-based recovery controls may be affected as well by attacks. As an example,
an AI-based managed backup system relying on an intrusion detection system (IDS) could 
see its IDS fooled by an adversarial machine learning (AML) attacker poisoning data in tandem
 with ransomware targeted to \added{encrypt} the backup database. 
 
\paragraph{AI-based targeted attacks.} 
As in other domains, \added{ attacks against AI systems can be generic (i.e., released in the wild) or targeted (due to being specially valuable for attackers because of the
stored information or the supported operations)}. In particular, attackers can use for such purpose AI-based attacks as it happens, e.g., with employing poisoned images to
overcome an image recognition system \citep{advclas} or enable side-channel attacks facilitating reverse engineering \citep{masure2020comprehensive}.
 Other attacks target AI systems \citep{sanyal2022towards} by e.g.\ injecting poisonous  information in the training data or taking advantage of function creep to abuse the system. 
  Moreover, cloud outsourcing in paradigms such as ML-as-a-Service implies third-party dependence and risks like model stealing or the existence of backdoors \citep{oliynyk2023know} in the outsourced AI models \citep{goldwasser2022planting}.
 
\section{SOLUTIONS TO NEW RISK ANALYSIS  \\ ISSUES IN SYSTEMS WITH AI COMPONENTS}
This section suggests solutions to the issues identified in Section 2 including a broad AI-related risk analysis framework (Section \ref{sec:cyber_risk_mgment_frame}).
When elsewhere available, particularly in Sections \ref{sec:AI_based_def} and \ref{sec:targeted_att},  we mainly summarise the literature
with relevant pointers. Section \ref{sec:example} details a case illustrating 
most of the issues raised here. 


\subsection{Impacts}
\label{sec:impacts}

\begin{figure}[h!]
\centering
\includegraphics[width=0.7\textwidth]{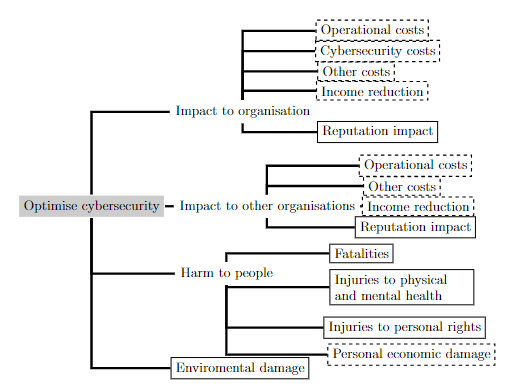}
\caption{Cybersecurity objectives (CSO) tree in CV20. Dashed box, monetary objective; solid line box, non-monetary.}\label{fig:tree}
\end{figure}

In order to adjust cybersecurity impact lists, in particular that in
 CV20, 
with those suggested in emerging AI risk management guidelines, we 
focus our discussion on the AIRMF, some of whose impacts are also 
present in the Act, the OECD AI Recommendation (\citeyear{oecd2019ai}) and the CRFM report, albeit with somewhat different names or structures.
\added{The CV20 list is structured as a cybersecurity objectives (CSO) tree (Figure \ref{fig:tree}) including objectives measurable in monetary terms ({\em operational costs, income reduction, cybersecurity costs, personal economic damage, other costs}) and others not directly measurable in such terms ({\em reputation, fatalities, physical and mental injuries,
injuries to personal rights, environmental damage}).} 

 The AIRMF focuses on the concept of \textit{trustworthiness}.
A trustworthy AI system should be 
  \textit{valid} and \textit{reliable},  \textit{safe},  \textit{secure} and \textit{resilient},  \textit{accountable} and \textit{transparent},  \textit{explainable} and \textit{interpretable}, 
  \textit{privacy-enhanced} and  \textit{fair}. Each of these features comprises additional sub-features. For example, validity entails {\em accuracy}, {\em reliability} and {\em robustness}. 
 Beyond these, the AIRMF identifies potential impacts associated with
 AI risks, such as \textit{harm to people}, \textit{organizations}, and  \textit{ecosystems}. As examples, it mentions \textit{harm to personal rights};  
 or \textit{impact over democratic participation}. The Act is more explicit identifying prohibited practices and high-risk AI systems, including their potential impacts. 
 For example, a forbidden functionality for an AI system
  would be {\em to employ subliminal techniques beyond a person's consciousness in order to materially distort his/her behaviour}.

\added{Interestingly, with appropriate modifications}, the CV20 \added{CSO} tree actually covers the 
  novel impacts identified in the AIRMF, the Act and the CRFM report. Such documents emphasise people's rights and physical risks in contrast to 
 cybersecurity frameworks from previous decades, in which those topics were less salient, 
  even absent. Moreover, the CV20 tree was constructed with decision support purposes in mind, integrating different types of objectives through multiattribute utilities and meeting standard requirements for decision support attributes (comprehensive, measurable, non-overlapping, relevant, unambiguous and understandable) \citep{keeney2005selecting}.
 Thus, we can actually use the CV20 tree to map those harms. 
 For instance, \textit{impacts on critical infrastructure} relate to the {\em impacts on other organizations} in the tree, a large-scale impact that harms 
 multiple organizations. Another example is {\em discrimination against groups}; 
  CV20 identified the UN Universal Human Rights Index Database as a major source to identify specific damages to personal and social rights,
   including discrimination against groups.

The AIRMF trustworthiness approach, implicit in the Act, would make for interesting candidates to expand our original tree. However, impacts in terms of 
AI trustworthiness features can also be translated into our objectives. 
As mentioned, trustworthiness could be interpreted by means of the CIA triad, taking into account additional points related to resiliency, accuracy, safety, and privacy (ISO, \citeyear{iso24028}) and  sustainability \citep{mcdaniel2022sustainability}.
  CV20 did not include them in the tree but rather integrated 
them into other objectives, like confidentiality aspects such as personal (personally identifiable information) or property rights (copyright, trademark, patent infringements), the organization's income or reputation (due to trade secret exposition)
or noncompliance with cybersecurity regulations.

All in all, as
Table \ref{table:impacts} reflects, following a similar approach, we are able 
to  map into the CV20 tree 
the impacts 
related to the different trustworthiness characteristics identified by AIRMF (and implicit in the EU and CRFM
proposals).

\begin{table}[h!]
\resizebox{\textwidth}{!}{%
\begin{tabular}{|ll|l|}
\hline
\multicolumn{2}{|l|}{\textbf{Trustworthiness principles and subprinciples}} &
  \multirow{2}{*}{\textbf{Map to CSO Tree \citep{couce2020assessing}}} \\ \cline{1-2}
\multicolumn{1}{|l|}{\textbf{Level 1}} &
  \textbf{Level 2} &
   \\ \hline
\multicolumn{1}{|l|}{\multirow{3}{*}{\textbf{Validity}\textsuperscript{a}}} &
  \begin{tabular}[c]{@{}l@{}}\textbf{Accuracy}\textsuperscript{a}: ``closeness of results of observations,\\ computations or estimates to true values or values\\   accepted as being true.'' \textsuperscript{e} \\ \end{tabular} &
  \multirow{3}{*}{\begin{tabular}[c]{@{}l@{}}Failing to achieve these characteristics involves a \\ degradation and maybe a malfunction or unavailability\\   of the AI system. These impacts may affect goals such \\ as operational costs, income reduction, other costs\\  (including non-compliance costs), reputational impact\\ or AI risk management costs.
  \end{tabular}} \\ \cline{2-2}
\multicolumn{1}{|l|}{} &
  \begin{tabular}[c]{@{}l@{}}\textbf{Reliability}\textsuperscript{a}: ``ability of an item to perform as\\ required, without failure, for a given time interval,\\ under given conditions.'' \textsuperscript{e} \end{tabular} &
   \\ \cline{2-2}
\multicolumn{1}{|l|}{} &
  \begin{tabular}[c]{@{}l@{}}\textbf{Robustness}\textsuperscript{a,b}: ``ability of a system to maintain its\\ level of performance under a variety of circumstances.'' \textsuperscript{e} \end{tabular} &
   \\ \hline
\multicolumn{2}{|l|}{\begin{tabular}[c]{@{}l@{}}\textbf{Safety}\textsuperscript{a,b}: ``property of a system such that it does not, under defined \\ conditions, lead to a state in which human life, health, property or \\ the environment is endangered.'' \textsuperscript{e} \\ {}\end{tabular}} &
  \begin{tabular}[c]{@{}l@{}}Impacts on this characteristic affect goals related to harm \\ to people. Specifically, fatalities or injuries to physical \\ and mental health. Indirectly, it may also create economic \\ damage to persons, the organization and third parties.\end{tabular} \\ \hline
\multicolumn{1}{|l|}{\multirow{3}{*}{\textbf{Fairness}\textsuperscript{a,b}}} &
  \begin{tabular}[c]{@{}l@{}}\textbf{Bias is managed}\textsuperscript{a,b}: ``systemic, computational, and\\ human {[}bias{]}, all of which can occur in the absence\\ of prejudice, partiality, or discriminatory intent.'' \textsuperscript{a} \end{tabular} &
  \begin{tabular}[c]{@{}l@{}}Impact on this characteristic may affect the reputation \\ impact goal (ethical degradation) and other costs\\ (non-compliance with AI regulation).\end{tabular} \\ \cline{2-3} 
\multicolumn{1}{|l|}{} &
  \begin{tabular}[c]{@{}l@{}}\textbf{Diversity}\textsuperscript{a,b}: ``reflect demographic diversity and broad\\ domain and user experience expertise.'' \textsuperscript{a} \end{tabular} &
  \multirow{2}{*}{\begin{tabular}[c]{@{}l@{}}Impact on these characteristics may affect goals related \\ to injuries to personal rights, personal economic damage \\ and even injuries to physical and mental health. \\ These consequences may also impact the organization \\ in terms of reputation (ethical degradation) and costs. \end{tabular}} \\ \cline{2-2}
\multicolumn{1}{|l|}{} &
  \begin{tabular}[c]{@{}l@{}}\textbf{Risk of discrimination is minimized}\textsuperscript{b,c}: ``based on\\any ground such as sex, race, color, ethnic or social\\ origin, genetic features, language, religion or\\ belief, political or any other opinion, membership\\ of a national minority, property, birth, disability,\\ age or sexual orientation.'' \textsuperscript{d} \end{tabular} &
   \\ \hline
\multicolumn{1}{|l|}{\multirow{2}{*}{\textbf{Security}\textsuperscript{a,b}}} &
  \begin{tabular}[c]{@{}l@{}}\textbf{Protection}\textsuperscript{a,b}: protocols to avoid, protect against, \\ respond to, or recover from attacks.'' \textsuperscript{a} \\ \end{tabular} &
  \multirow{2}{*}{\begin{tabular}[c]{@{}l@{}}These characteristics are represented by the risk \\  management controls, affecting also the risk \\ management costs goal.\end{tabular}} \\ \cline{2-2}
\multicolumn{1}{|l|}{} &
  \begin{tabular}[c]{@{}l@{}}\textbf{Resilience}\textsuperscript{a}: ``ability to return to normal\\ function after an unexpected adverse event.'' \textsuperscript{a} \end{tabular} &
   \\ \hline
\multicolumn{1}{|l|}{\multirow{3}{*}{\textbf{AI Governance}}} &
  \begin{tabular}[c]{@{}l@{}}\textbf{Transparency}\textsuperscript{a,b}: ``the extent to which information is\\ available to individuals about an AI system, if they\\ are interacting – or even aware that they are\\ interacting – with such a system.'' \textsuperscript{a} \end{tabular} &
  \multirow{3}{*}{\begin{tabular}[c]{@{}l@{}}Failing to achieve these characteristics involves a\\ reputational impact (ethical degradation) and may \\ include  other costs  for non-compliance.  \end{tabular}} \\ \cline{2-2}
\multicolumn{1}{|l|}{} &
  \begin{tabular}[c]{@{}l@{}}\textbf{Accountability}\textsuperscript{a,b,c}: ``expectations of the responsible\\ party in the event that a risky outcome is realized.'' \textsuperscript{a} \end{tabular} &
   \\ \cline{2-2}
\multicolumn{1}{|l|}{} &
  \begin{tabular}[c]{@{}l@{}}\textbf{Human oversight}\textsuperscript{c}: ``natural persons can effectively\\ oversee the AI system during the period in which\\ the AI system is in use.'' \textsuperscript{c} \end{tabular} &
   \\ \hline
\multicolumn{1}{|l|}{\multirow{2}{*}{\textbf{Understandability}}} &
  \begin{tabular}[c]{@{}l@{}}\textbf{Explainability}\textsuperscript{a,b}: ``representation of the mechanisms\\ underlying an algorithm’s operation.'' \textsuperscript{a} \end{tabular} &
  \multirow{3}{*}{\begin{tabular}[c]{@{}l@{}}Failing to achieve these characteristics involves a\\ degradation and maybe a malfunction or unavailability\\   of the function performed by the AI system. Impacts \\ on these may affect goals of the organization such as \\  operational costs, income reduction, other costs  \\  (including non-compliance costs), reputational impact\\ or risk management costs. \end{tabular}} \\ \cline{2-2}
\multicolumn{1}{|l|}{} &
  \begin{tabular}[c]{@{}l@{}}\textbf{Interpretability}\textsuperscript{a,b}: ``the meaning of AI systems’ output\\ in the context of its designed functional purpose.'' \textsuperscript{a} \end{tabular} &
   \\ \cline{1-2}
\multicolumn{1}{|l|}{\multirow{2}{*}{\textbf{Data governance}\textsuperscript{b}}} &
  \begin{tabular}[c]{@{}l@{}}\textbf{Data management}\textsuperscript{c}: ``data collection, data analysis,\\ data labelling, data storage, data filtration, data\\ mining, data aggregation, data retention and any\\ other operation regarding the data.'' \textsuperscript{c} \end{tabular} &
   \\ \cline{2-3} 
\multicolumn{1}{|l|}{} &
  \begin{tabular}[c]{@{}l@{}}\textbf{Privacy}\textsuperscript{a,b}: ``safeguard human autonomy, identity\\ and dignity.'' \textsuperscript{a} \end{tabular} &
  \begin{tabular}[c]{@{}l@{}}Privacy affects the goal of injuries to personal rights \\ and may include other costs for non-compliance. \end{tabular} \\ \hline
\end{tabular}%
}
\scriptsize	{ References: \newline \textsuperscript{a} NIST AI Risk Management Framework \citep{nist2022ai}, \newline \textsuperscript{b} CRFM's On the Opportunities and Risks of Foundation Models \citep{Stanford2021FoundationModels}, \newline \textsuperscript{c} EU AI Act \citep{commissie2021proposal}, \newline \textsuperscript{d} Article 21 of the EU Charter of Fundamental Rights (used in \textsuperscript{c}) \citep{commissie2012charter}, \newline \textsuperscript{e} ISO/IEC TS 5723:2002 (used in \textsuperscript{a}) \citep{iso5723}.}\\
\caption{Mapping trustworthiness features into cybersecurity objectives. Indices in terms identify references addressing the principle. Indices in definitions identify source.}
\label{table:impacts}
\end{table}

\subsection{A cybersecurity risk analysis framework for \\
systems with AI components
}
\label{sec:cyber_risk_mgment_frame}

The basic structure in the originating  framework \citep{rios2021adversarial} essentially used
a single block to conceptualize a cyber organization. Given the relevance of AI 
components as reflected in Section 2 challenges, here 
we structure organizations with finer granularity in terms
of blocks or components, which could refer to hardware, software or hardware-software
elements.  This section provides a  
 broad description of the  
cyber risk analysis framework and then  
reflect on the AI ingredients (assets and defenses) included.

\subsubsection{\textit{\mdseries {Structure}}}

\begin{figure}[h!]
\includegraphics[width=0.6\textwidth]{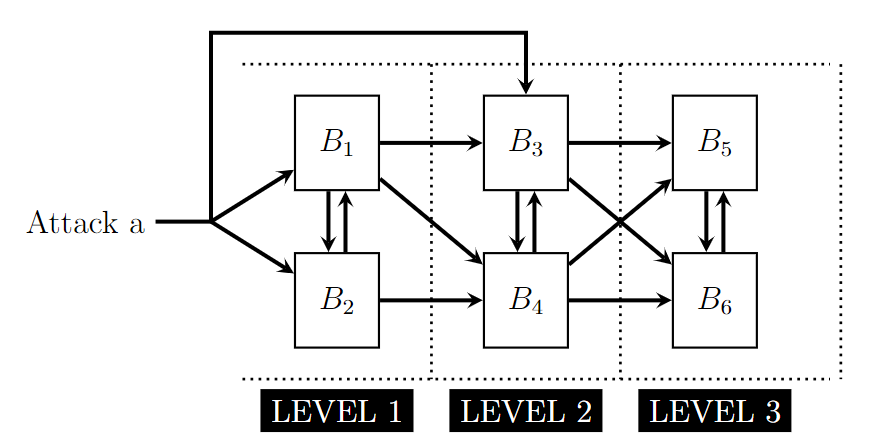}
\centering
\caption{Cyber organization structured according to 6 blocks in three levels.}
\label{fig:network_3}
\end{figure}

Consider a cyber organization structured according to access, security and criticality levels, much as in the Purdue Model for Industrial Control Systems \citep{williams1994purdue}, 
 represented through blocks in levels as Figure \ref{fig:network_3} exemplifies. 
 Let $G$ be the underlying graph with nodes representing blocks
 and links, flows through which an attack could enter or be transferred.
  $G$ might be cyclic at a given level. This representation provides a flexible method to model cyber systems including AI components.

\paragraph{Example 1.}
Consider an organization structured according to six blocks $B_1, B_2, $ $..., B_6$, distributed along three levels (Figure \ref{fig:network_3}). 

\noindent According to it,  a certain type of attack might enter through $B_1$, $B_2$
(level-1 blocks),  or $B_3$ (level-2 block).   \hfill $\triangle$\\

\noindent  Formally, the parameters describing a system will be as follows,
where, for the moment, we assume there is just one type $a$ of attack:
\begin{itemize}
    \item  The number $k$ of levels.
    \item  The number and name of blocks within each level
    \[
\begin{blockarray}{cccccc}
\begin{block}{cc[ccc]c}
 & \text{Level 1} & B_{1},B_{2}& \cdots & B_{i_{1}} & i_1 \text{ blocks}, \\
 & \text{Level 2} & B_{i_{1}+1} & \cdots & B_{i_{2}} & i_{2} - i_{1} \text{ blocks}, \\
   & \vdots & \vdots & \vdots & \vdots \\
 & \text{Level $k$} & B_{i_{k-1}+1} & \cdots & B_{i_{k}} & i_{k} - i_{k-1} \text{ blocks}. \\
\end{block}
\end{blockarray}
 \] 
 Blocks within a level 
are numbered in an ascending manner (if a block precedes  
another one, but not otherwise,
its index is smaller). Importantly some of the blocks could refer to AI-based assets, as we later reflect. 
\end{itemize}
\begin{itemize}
    \item  The entrance probabilities, that is the probabilities  
     with which the attack 
    will attempt to enter through various  blocks (some of them  
     could be 0), 
 \[
\begin{blockarray}{cccccccc}
\begin{block}{c[cccccc]c}
P_1 =  & p_1, & p_2, & \cdots & ,\quad p_j,  & \cdots & p_{i_k}  & \text{of the
 attack  entering just through 
 block $B_j$,}\\ 
\end{block}
\end{blockarray}
\] 
 $
P_2 = \left[ p_{12}, p_{13}, ..., p_{1i_{k}}, p_{23}, \cdots , p_{2i_{k}},...,p_{ij},...,
   p_{{i_k} -1, {i_k} }  \right] $ just through 
 blocks $B_i$ and $B_j$, simultaneously
\[... \]
\[
\begin{blockarray}{ccc}
\begin{block}{c[c]c}
P_{i_k} = & p_{12\cdots i_{k}} & \text{\,\, 
simultaneously through all the blocks,} \\
\end{block}
\end{blockarray}
\]
with $\sum _i p_i +\sum_{i<j} p_{ij}+... + p_{12\cdots i_{k}}=1 $. $P$ will designate the union of $P_1 ,  P_2 , \ldots , P_{i_k}.$ 
\end{itemize}

\noindent 

\begin{itemize}
    \item $ Q = [q_i]$, the probability of not protecting (PNP) block $B_i$ from such type
    of external attack, if attacked. Obviously,  the probability of protecting it is $1- q_i$.
    \end{itemize}
    
    \begin{itemize}
    \item  For level $s$ ($s = 1, 2, \ldots, k$), $Q ^s = [q_{ij}]$ contains the PNPs block $B_i$ from block $B_j$ (where $j$ is at level $s$ or ($s-1$)) if there is an attack following an information transfer from $B_j$ into $B_i$. 
\end{itemize}
The $q$ probabilities would refer to the type of defense considered,
in particular to whether they incorporate AI-based defense systems, as 
  section \ref{sec:AI_based_def} specifies.

\paragraph{\bf Example 1 (cont).}
 For the system in Figure \ref{fig:network_3}, we would have 
\begin{itemize}
\item Blocks $B_i, i\in\{1,2,3,4,5,6\}$ distributed in levels $s\in \{1,2,3\}$.  
\item $p_1$, $p_2$, $p_3$: the probabilities of accessing the system just through 
one of the blocks $B_{1}$, $B_{2}$, $B_3$, (additionally, $p_4$ =  $p_5$ = $p_6$ = 0);
   $p_{12}$, $p_{13}$, $p_{23}$: the probabilities of accessing the system 
     through just the pairs of 
    blocks ($B_{1}$, $B_2$), ($B_{1}$, $B_3$), ($B_{2}$, $B_{3}$), respectively (the remaining $p_{ij}$ would be 0); 
    $p_{123}$: the probability of accessing simultaneously through $B_1$, $B_2$ and $B_3$ (the remaining $p_{ijk}$ would be 0).
 As mentioned, $\sum_{i=1}^3 p_i + \sum _{ij\in 
\{12,13,23\} } p_{ij} + p_{123}= 1$. 

\item $q_1$, $q_2$, $q_3$ are the block PNPs if they are attacked externally. $q_{ij}$, $(i,j)$ $\in$ $\{$$(1,2),$ $(2,1),$ $(3,1),$ $(3,4),$ $(4,1),$$(4,2),$ $(4,3),$ $(5,3),$ $(5,4),$ $(6,3),$$(6,4)$ $\}$ represent the PNPs of $B_i$ from block $B_j$. \hfill $\triangle$

\end{itemize}

\subsubsection{\textcolor{black}{A risk analysis pipeline for AI based systems}}
\label{sec:attack_simulation}
Using the above structure, we provide
algorithms to simulate the propagation of 
an attack and its eventual impacts on an AI based system, and employ them to 
perform risk analysis. 

\paragraph{Attack transit simulation}
\label{sec:transit_sim}

Algorithm \ref{alg:bis2} simulates the transit of an attack within an organization
using the above structure. It outputs 
an indicator $I_i$ for each block $B_i$, so that 
 $I_i = 1 $ (0) if the block has (not) been successfully attacked.  
   $N^{'}$ designates a sufficiently big number (interpretable 
   as the expected number of transits of the attack before detected or before completing
   its function); if necessary, it may be replaced by parameter $N^{'}_{jh}$ referring to the transits between blocks $j$ and $h$. $N^{'}$ or the $N^{'}_{jh}$ would be generated randomly from properly selected distributions, illustrated in the case study.
 Observe that if the graph is acyclic, we would make
   $N^{'}=1$, there being just one transition.
\begin{algorithm}[h!]
\caption{Attack transit simulation in a general facility} 
\label{alg:bis2}
\textbf{Input:} $G$, $P$, $Q$, $Q^s$ 
\begin{algorithmic}[1]
\State $I_j$ = 0, $\forall$ block $B_j$
\State Given $P$,  simulate the attacked entries and assign $I_j = 1 $ to them. \label{alg1:step2}
\State Given $Q$, assign  $I_j = 0$ to non-successfully attacked entries.

\If{all $I_j = 0$} 
    \State Stop
  \Else 
     \For{$h \in \{1, \ldots , k \}$} \Comment{Loop in level-depth}
     \For{$u \in \{1, \ldots , N^{'} \}$}
          \For{$j \in \{i_{h} +1, \ldots , i_{h+1} -1 \}$}
\If{$I_j = 0$} 
\State Next $j$
\Else 
\For{each successor $h$ of block $j$}
\If{$I_h =0 $}
\If{successor $h$ successfully attacked from block $j$}
\State $I_h =1$
\EndIf
\EndIf
\EndFor
\EndIf
\EndFor
\EndFor
\EndFor
\EndIf
\State \textbf{Return:} $I_j$, $\forall$ block $j$
\end{algorithmic}
\end{algorithm}

\noindent Key points in the algorithm specification are:

\begin{itemize}
    \item  Step 2, simulating attack entry points.
    For a given facility with entrance probabilities 
    $P$, we generate one sample from a multinomial with such probabilities.
     A Dirichlet distribution could be used to generate $P$,
    given the uncertainty about it.\footnote{This distribution, as the others considered in this manuscript, would evolve as data accumulates.} In principle, this requires up to $2^n - 1$ probabilities, with $n$ the number of identified single entry points, a shortcoming from a storage point of view. A realistic way to mitigate this is to assume that, with a certain probability, say generated from a beta distribution, the attack is generic affecting all entry point blocks, whereas the remaining probability is allocated equally to targeted attacks on the identified single entry points.

\item Steps 3 and 15, simulate the success of an attack to a block, based 
on the PNP $q$ of such block from an attack.  Typically $q$ would be generated from beta distributions.  Section 3.3 discusses how to assess $q$,
in particular for AI-based blocks.
 \end{itemize}

\paragraph{Simulation and aggregation of impacts}
\label{sec:sim_agg_imp}
 Algorithm \ref{alg:bis2} outputs a configuration $(I_1, \ldots, I_{i_k})$ indicating the blocks affected by the attack. We next focus on the associated impacts
 simulating them at block or system level, depending on the scenario modeled: 
some impacts will be {\em global} (one impact for the whole system),
 whereas others will be {\em separable} (one impact per block).
Separable impacts are aggregated through a rule $b$. Finally, multiple impacts are aggregated with a rule $g$. \\

\noindent \textbf{Example 1 (cont).} Recalling Section 3.1, suppose the incumbent type of attack may cause the following impacts: { \em financial} ($l_1$), {\em equipment damage} ($l_2$), 
and {\em downtime} ($l_3$). $l_1$ is global, whereas $l_2$ and $l_3$ are separable. Typical aggregating details would be: 
\begin{itemize}
\item  $l_{2}$. If $l_{2_1}, \ldots, l_{2_{i_k}}$ are the corresponding equipment replacement costs (0 if no replacement required), a plausible aggregation would be  $l_2 = \sum_{j} l_{2j}$. 
\item  $l_{3}$. For aggregation purposes, multiply it by the cost of an hour of 
downtime for the corresponding installation.  If $l_{31}, \ldots, l_{3_{i_k}}$ are the downtimes, we would expect $l_3 = \max  (l_{31}, \ldots ,l_{3_{i_k}}) \times d$,where $d$ is the unit downtime cost. 
\end{itemize}
Once the impacts are computed, we aggregate them through a rule $l  = g(l_{1}, l_{2}, l_{3})$ for example, using a multi-attribute weighted value function
$g(l_{1}, l_{2}, l_{3}) = \sum_{i=1}^{3}{w_i\times l_i}$, where $w_i$ weighs the importance of the $i$-th impact, and even transform it through, say, a risk-averse utility function \citep{GonzalezOrtega2018}.
$\hfill \triangle$

\vspace{0.3cm}

\noindent At schematic level, simulating from the impact distribution based on $(I_{1}, \ldots, I_{i_k})$ would run as in Algorithm \ref{alg:MCBI}, with the 
first $L$ impacts assumed to be local and the remaining $R-L$, global. The $j$-th impact,
  $j=1,...,L$ would have its corresponding aggregation rule $b_j$.

\begin{algorithm}[h!]
\caption{Simulation from impact distribution} 
\label{alg:MCBI}
\hspace*{\algorithmicindent} \textbf{Input:} $(I_{1}, \ldots, I_{i_k})$ 
\hspace*{\algorithmicindent} 
\begin{algorithmic}[1]
\State $l_1, l_2, ..., l_R$ = 0

\If{$\max(I_1, \ldots, I_{i_k})$ = 0} 
    \State $l = 0$
  \Else 
    \For{$r \in \{L+1, \ldots, R\}$}
    \State Generate $l_{r}$
    \EndFor
     \For{$h \in \{1, \ldots , i_k \}$}
       \If{$I_h $ = 1}
        \For{$j  \in \{1, \ldots, L\}$}
        \State Generate $l _{j}^h$
        \EndFor
     \EndIf
      \EndFor
    \For{$j  \in \{1, \ldots, L\}$}
     \State Aggregate $l_j ^h $ with rule $b_j $ obtaining $l^{'}_j$
\EndFor
\State Aggregate the $l^{'}_j$'s and $l_r$'s with rule $g$ obtaining $l$
\EndIf
\State \textbf{Return:} $l$
\end{algorithmic}
\end{algorithm}

\noindent This procedure generates 
a sample of impacts that serves as a basis to build 
the loss curves required for risk management.

\paragraph{General scheme for attack simulation}
Algorithm \ref{alg:SGAF} simulates the consequences associated with a specific attack of a given type $a$. For such type, a stochastic process $\Lambda$ generating attacks in the required 
  risk analysis planning period would be called upon. Then, its 
routine would be invoked for each generated attack. There would be a stochastic process for each type of relevant attack with its own parameters,  and the corresponding generating routine defining its arrival process. $M$ would be the sample size required to estimate the different quantities (expected 
   costs, utilities) used below to the desired precision.

\begin{algorithm}[h!]
\caption{\textcolor{black}{Attack simulation in a general facility}} 
\label{alg:SGAF}
\hspace*{\algorithmicindent} \textbf{Input:} \textcolor{black}{$G$, $\Lambda$, distributions for $P$,$Q$,$Q^{s}$}
\hspace*{\algorithmicindent} 
\begin{algorithmic}[1]
\For{\textcolor{black}{$j \in \{1, \ldots, M\}$}}
\State \textcolor{black}{Generate number $N$ of attacks using $\Lambda$}
\For{\textcolor{black}{$i \in \{1, \ldots, N\}$}}
\State \textcolor{black}{Generate $P$, $Q$, $Q^{s}$}
\State \textcolor{black}{Simulate attack transit using $P$,$Q$, $Q^s$\Comment{Algorithm \ref{alg:bis2}}}
\State \textcolor{black}{Simulate impact $l_i$ from attack \Comment{Algorithm \ref{alg:MCBI}}}
\EndFor
\State \textcolor{black}{Aggregate $l^j = \sum_{i=1}^{N}l_i$}
\EndFor
\State \textcolor{black}{\textbf{Return:}} \textcolor{black}{$\{l^j \}_{j=1}^{M}$}
\end{algorithmic}
\end{algorithm}

\paragraph{Risk assessment}

 Once we have obtained $M$ samples from the impact distribution, we can estimate

    \subparagraph{Probabilities.} The $I_j$ outputs are used to estimate the 
     attack probabilities on various blocks by just counting the number of 1's appearing for each block and dividing by the number $M$ of simulation iterations. This facilitates assessing the vulnerability of each block.

    \subparagraph{Losses.} The output $\{ l^j \}_{j=1}^{M}$ is fit to  a mixed-type
    distribution.
 Typically, it will have a mass point at $l=0$ and we would fit a density to the positive observations, through   
    a mixture of gamma distributions, for reasons outlined in \cite{wiper}. The 
    estimated model can be summarized through the probability of the zero part and the moments of the positive part and quantities like the VaR or CVaR. If considered sufficiently 
    risky, we would proceed with risk management.
    
    Beyond the global loss 
    at system level, it will typically be interesting to analyze component losses 
     at block level for the separable impacts to assess their contributions.

\paragraph{Risk management}
\label{sec:cyber_mitigation_selection}
We next present an approach to risk management (RM) linked with the methodology introduced and addressing these issues: 
 includes the costs of the security portfolio  
     in the assessment; includes cyber insurance 
     adoption as part of the risk management process; 
     caters for taking into account the
     organization's risk attitude; introduces constraints over portfolios; 
     takes into account the uncertainty in portfolio evaluation due to sampling.
 Let us formulate first the problem and then introduce a generic algorithm to solve it. 

\subparagraph{RM formulation}

Cyber mitigation portfolios will be characterized by a vector 
${\bf c}=(c_1,....,c_m)$ with $c_i\in \{0,1,2\}$ indicating whether  
the $i$-th control: is (not) included in the portfolio when $c_i=1$
($c_i=0$); is already implemented when $c_i=2$. Some of the controls might correspond to AI-based
components; one of them could refer to a (cyber)insurance product. To wit, an
 organization will typically 
have a few mitigations already implemented, say the first $r$,
that is $c_1=\ldots = c_r = 2$. 
Besides, for compliance reasons, some of the mitigations 
would be enforced, say from $r+1$ to $s$,
 so that the initial configuration is designated 
${\bf c}^{*0}=(2,...,2,1,...,1,0,...,0)$.
The aim is to decide which of the controls 
$(c_{s+1},...,c_{m})$ should be implemented additionally. \textcolor{black}{For this, given a proposed configuration ${\bf c}$,  we shall have the corresponding entrance 
 probabilities $ P({\bf c}) $ and PNPs  $\big(Q({\bf c}), Q^s ({\bf c})\big)$}.
 A simulation routine as in 
Algorithm 3 would provide a sample
$\{ l ({\bf c})^j \}_{j =1}^ M$
of the loss if portfolio $\textbf{c}$ is implemented. 

Thus, the problem we aim to solve is 
\begin{equation}\label{kkparis}
 \max  \quad E ( u ({\bf c})  ) \,\,\, 
 \end{equation}
\[ {\rm s.t.} \,\, c_i \in \{0,1,2\}; c_i=2, i\leq r, c_i=1, r+1\leq i \leq s, \]
\[ h({\bf c}) \leq 0 ,\] 
 where $ u$ is the utility function modeling preferences and risk attitudes, and $h({\bf c})\leq 0$ designates relevant constraints.

Concerning  the objective function, we first integrate the portfolio 
$cost({\bf c})$ in the loss, which adopts the form $ l ( {\bf c})+cost({\bf c})$, where $cost({\bf c})$  is typically based on the sum of the included mitigation costs.
Next, we adopt a constant
absolute risk averse (CARA) utility function \citep{GonzalezOrtega2018} whose 
form\footnote{ We aim at minimizing costs, therefore maximizing -costs.} is
$u({\bf c})= 
1 - \exp (\rho ( l ({\bf c}) + cost ({\bf c}) )),$
with $\rho$  being the risk aversion coefficient.
 Two main types of constraints would refer to:  \\
\noindent - \textit{Budget.} Typically, there would be a constraint indicating the maximum budget $C$ available for cyber mitigation, 
\[ \sum _{i = \textcolor{black}{1}} ^m cost (c_i) \leq C,\] where $cost (c_i)$ is the cost of the $i$-th mitigation. Splits between maintenance and implementation costs could be introduced. If $ic$ represents the maximum implementation budget, the constraint would be  
\begin{equation*}
\sum _{i=r+1}^m c_i \times icost (i) \leq ic .
\end{equation*}
Similarly, for $mc$ representing the maximum maintenance budget, the constraint
would be 
\begin{equation*}
\sum _{i=1}^{\textcolor{black}{m}}
\sign(c_i) \times mcost(i) \leq mc.
\end{equation*}

\noindent - \textit{Compliance} with laws, standards, and frameworks may enforce certain mitigations to be implemented, as mentioned above, entailing a reduction in the  budgets available.

\subparagraph{Implementation}\label{section4}
A first strategy would be to search the space of portfolios, find out their corresponding probabilities
$P({\bf c})$, $Q({\bf c})$, and $ Q^s ({\bf c})$,
  obtain a sample from the loss $l({\bf c})$ to estimate the expected utility of portfolio \textbf{c}, and optimize it,
  with the aid of a discrete stochastic optimization method, see \cite{powell2019unified}.
 This may become too cumbersome
 computationally when the set of feasible portfolios is large. 

Alternatively, based on the typical form 
of the loss curves, Figure \ref{kkjlo},
we would assume that the loss could be modeled 
 as a mixture\footnote{A similar procedure would 
 be followed when the gamma distribution in (\ref{oil}) is replaced 
 by a mixture of gamma distributions.} 
\begin{equation}\label{oil}
\mathcal{L}({\bf c}) \sim  s({\bf c}) I_{0} + \big(1 - s({\bf c})\big) Gamma\big(a, t({\bf c})\big)
\end{equation}
where $s({\bf c})$ represents the probability of no loss if 
${\bf c}$ is the implemented portfolio and
$t({\bf c})$ adopts a specific parametric form to adapt to the shape of positive losses. For 
the $i$-th mitigation, we have its implementation $(icost(i))$ and maintenance
 $(mcost(i))$ costs and parameters describing its effectiveness ($\alpha_i$, $\beta_i$).
 The parametric forms that we adopt for $s({\bf c})$
and $t({\bf c})$ are

\begin{equation*}
s({\bf c}) = 1 - s_{0}\exp\Big(-\Big(\sum_{i=1}^{m} \alpha_{i} \times \sign(c_{i})\Big)\Big),
\,\,\, 
t ({\bf c}) = t_0 + \sum_{i=1}^{m} \beta_{i} \times \sign (c_{i}),
\end{equation*}
 suggesting diminishing returns in cybersecurity investments,
whereas the corresponding costs will be 
\[ cost ({\bf c}) = \sum _{i=r+1}^m c_i \times icost (i) + \sum _{i=1}^m
\sign(c_i) \times mcost(i).\]

\noindent The objective is therefore to provide $M$ samples $(cost_1 , \ldots, cost_M  )$ from the cost  and, correspondingly, from
the utility $( u_1, \ldots, u_ M )$. 

We include now the algorithms that implement the RM setup. First,  
Algorithm \ref{alg:generate_sample} estimates, for a given portfolio  \textbf{c},
its expected utility, updating the parameters given the implemented controls,
in particular, including its implementation ($ticost$) and maintenance ($tmcost$) costs. 

\begin{algorithm}[H]
\caption{\text{Compute expected utility}} 
\label{alg:generate_sample}
\textbf{Input:} ${\bf c}$, $s_0$, $t_0$, $\alpha$, $\beta$, $a$, $M$,
$\rho$, {\em mcost}, {\em icost}
\begin{algorithmic}[1]
\State Compute $tmcost$ and $ticost$
\State $ancost = tmcost + ticost$
\State $util=0$
\For{$ i \in \{1, \ldots , M \}$}
\If{$(c(i) = 1)$ OR $(c(i) = 2)$}
\State $s_0 = s_0 \times \exp\big(-\alpha(i)\big)$ 
\State $t_{0} = t_{0} + \beta(i)$
\EndIf 
\EndFor
\State $s = 1 - s_0$
\State $t = t_0$

\For{ $i = 1, \ldots, M$ } \label{alg4:initial_loop}
\State Generate $u\sim U(0,1)$
\If{$u<s$}
\State $cost(i) = 0$
\Else
\State $cost(i)\sim Gamma(a,t)$
\EndIf
\State $cost(i)=cost(i)+ancost$
\State $util=util +  (1- \exp\big(\rho \times cost(i)\big) ) $ \label{alg4:final_loop}
\EndFor
\State \textbf{Return:}  $util/M$ 
\end{algorithmic}
\end{algorithm}

\noindent  Algorithm \ref{alg:global} integrates the previous pieces.
Given the portfolio, its feasibility is first assessed. If feasible but not optimal,
the portfolio is updated, 
where Algorithm * designates a generic routine proposing 
a portfolio update 
for optimization purposes, see \cite{powell2019unified} for pointers including 
simulated annealing.

\noindent 

\begin{algorithm}[H]
\caption{Global scheme} 
\label{alg:global}
\hspace*{\algorithmicindent} 
\begin{algorithmic}[1]
\State Current ${\bf c}$
\State Compute portfolio costs
\If {''Infeasible Portfolio''}
\State Stop
\EndIf
\While{not convergent}
\State Compute portfolio costs
\If{``Infeasible Portfolio''}
\State Stop
\Else
\State Update $s_0$ and $t_0$ \Comment{Algorithm} \ref{alg:generate_sample}
\State Estimate expected utility of portfolio  
\Comment{Algorithm \ref{alg:generate_sample}}
\State Update {\bf c} \Comment{Algorithm *}
\EndIf
\EndWhile
\end{algorithmic}
\end{algorithm}

 \subsection{AI based defenses}
\label{sec:AI_based_def}

The PNP parameters $q$  
assessing the non-protection of blocks against attacks are key model inputs. They are characteristic of each defense type 
against a particular attack type in a given environment. We may have data  
 and/or use expert judgment \citep{hanea2021expert} to estimate and 
  update them in the light of data. 

When handling AI systems that are safety components of products,  as with content filters or \textcolor{black}{computer vision}
systems \citep{COMITER2019}, a major information source to assess 
the parameters are their {\em security evaluation curves}.
This is
part of the relatively recent domain of AML, see \cite{BIGGIO2018317,vorobeichikantar,advclas,gallego} for reviews.
These curves depict the probability $(1-q)$ of succeeding in protecting an ML algorithm 
from an attack of a certain type and intensity, 
  given the AI-based 
defense implemented. \textcolor{black}{ Hence, given a type of attack, its intensity 
and the chosen defense, the parameter $1-q$ would be estimated with 
the corresponding curve, from which the PNP would be deduced.} 

\begin{figure}[!h]
\centering
\includegraphics[width=.65\linewidth]{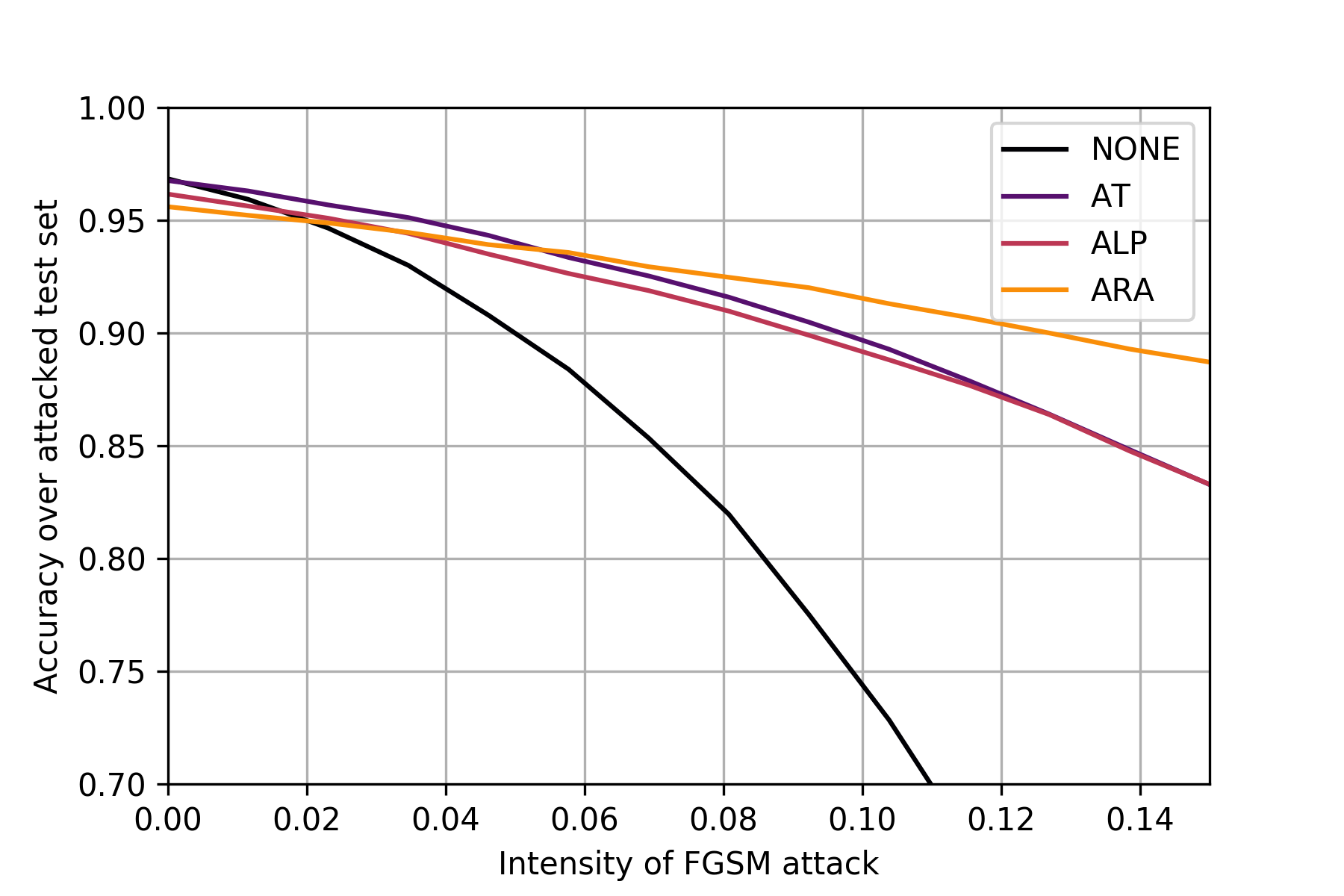}
  \caption{Security evaluation curve of a deep network for MNIST data under four defense mechanisms
(NONE, AT, ALP, ARA) against FGSM attack. From \cite{gallego}.}
  \label{fig:sub1a}
\end{figure}

As an example, Figure \ref{fig:sub1a} illustrates the security evaluation curves 
 of four ML defenses (none, adversarial training (AT), 
adversarial logit pairing (ALP), adversarial risk analysis (ARA)) over
a classifier used in a computer vision task against 
  a   
fast gradient sign method (FGSM) \citep{szegedy2013intriguin} attack of increasing intensity. 
 For a 0.06-intensity attack, the estimated expected accuracy 
 in the task would be around 0.93 (with ARA, AT and ALP defenses) and 
 about 0.87 with no defense and $\hat{q}$ would be, respectively,  
 0.07 and 0.13.  
 We would complete the assessment acknowledging the uncertainty about the corresponding $\hat{q}$
with a beta distribution with, e.g., such value as mode.

 \subsection{Targeted attacks}
 \label{sec:targeted_att}
  As Section 3.2 described, other key inputs are the attack probabilities $p_i$ to blocks $B_i$ from external sources for a given type of attack 
 as well as the stochastic process generating such type of attacks. As in 
 Section 3.3, we may have access to data and/or expert judgment which allows us to model and 
 estimate the corresponding probabilities and stochastic processes parameters. Special emphasis should be placed on targeted attacks, since they entail strategic calculations which further complicate the assessment 
of attack uncertainties.
For this, we appeal to ARA tools, overviewed in \cite{banks2022adversarial}, which require
\begin{enumerate}
\item Formulating the defender problem, that is, selecting a portfolio minimizing the potential impact of attackers' actions on our system. The uncertainties include whether our system, taking into account the implemented portfolio, will be targeted or if, alternatively, another system will be the focus. We also assess the uncertainties 
about potential attack entry points. 

    \item Formulating the attackers' decision problems, followed by 
  gathering  (typically partial) information  about the attackers' beliefs and preferences, leading to his random probabilities and random utilities from the defender's perspective.

\item Simulating from the attackers' problems to assess the attack probabilities and processes over the system of interest. This entails sampling from 
their  utilities and probabilities, and assessing their optimal attacks (type,
target and, if our system is the target, entry points). This is iterated 
as much as requested (and computationally feasible).
Based on this sample we assess the probability of receiving an attack at various entry 
points. 

\item The inclusion of such forecasts in the defender's problem (step 1), and its solution
to select the optimal protection portfolio. 
\end{enumerate}
 Note that whether targeted attacks are automated through AI systems or manually would 
 make little difference in modeling terms and we would just need to assess if the capabilities of the attacker enable 
   it to implement more sophisticated AI attacks or whether it has sufficient  resources to hire an AI-based attacking platform. In this sense, the seemingly ever-increasing availability 
   of such platforms, as 
   in AI-based Crime-As-A-Service \citep{kaloudi2020ai},    
   increases the importance of 
   developing this type of models. 

 Thus, a key issue in relation to targeted attacks is whether the system under study would be the target of interest
 of a specific attacker for a given attack. The framework sketched may be used to answer such type of questions based on the principle
 that a system  would be targeted if the
 attacker derives higher expected utility from attacking to it  
 than to its competitors (and from not attacking). As a byproduct, we also  determine the entry point probabilities.  To wit, the scenario is modeled as a sequential defend-attack game \citep{rios2012adversarial}, where we posit that attackers operate as expected utility maximizers and that they possess knowledge of the implemented portfolio for system protection before initiating any action.
Assume that the attacker can perpetrate the following actions compiled in set $\mathcal{A}$:

\begin{itemize}
\item[-] $a_{j,k}^1$, $j\in \{1,\ldots, J\}, k \in \{1, \ldots, K\}$:   Attack of type $j$ targeting the system's $k$-th (possibly multiple) entry,  with $J$ the number of attack types available, and $K$, the number of relevant entry combinations. Our assumption is that the attacker will select only one type of attack at any given moment (as opposed to employing multiple types concurrently). 

\item[-] $a^i_{j}, j\in\{1,\ldots, J\} , i\in\{2,\ldots, n\}$: Attack of type $j$ targeting another system $i$, with $\{2,\ldots, n\}$ encompassing all 
other targets  that the potential attacker may choose.

\end{itemize}

\noindent The expected utility $\psi_A$ when the attacker opts to target via 
$a_{(j,k)}^1$ our system given portfolio $\textbf{c}$  is defined through 
\begin{equation*}
\psi_A(a_{(j,k)}^1, \textbf{c}) = \\ p_{A}(Y=1|a_{(j,k)}^1, \textbf{c})u_{A}(Y=1, a_{(j,k)}^1, \textbf{c})+p_{A}(Y=0|a_{(j,k)}^1, \textbf{c})u_{A}(Y=0, a_{(j,k)}^1, \textbf{c}),
\end{equation*}
\noindent where $p_{A}(Y=1|a_{(j,k)}^1, \textbf{c})$ denotes the probability that the attack successfully penetrates the system, given  $\textbf{c}$ and the attack $j$ targeting the $k$-th entry, and $u_{A}(Y=1, a_{(j,k)}^1, \textbf{c})$  designates the utility of a successful attack $j$ targeting block $k$, given $\textbf{c}$, for the attacker. Given the limited knowledge of the attacker's utilities $U_A$ and probabilities $P_A$, we opt for a Bayesian approach leading to random utilities and probabilities, which make up for the attacker random expected utility   
\begin{equation*}
\Psi_A(a_{(j,k)}^1, \textbf{c}) = \\ P_{A}(Y=1|a_{(j,k)}^1, \textbf{c})U_{A}(Y=1, a_{(j,k)}^1, \textbf{c})+P_{A}(Y=0|a_{(j,k)}^1, \textbf{c})U_{A}(Y=0, a_{(j,k)}^1, \textbf{c}).
\end{equation*}
Similarly, from the defender's perspective, the random expected utility when the attacker
 targets the $i$-th system with attack $a_j$ is\footnote{ Note that we do not account for 
 changes in defenses of other systems, recognising a lack of detailed knowledge about the security status of the other systems.}  
\begin{equation*}
\Psi_A(a^i_j ) = P_{A}(Y=1|a^i_j)U_{A}(Y=1, a^i_j) + P_{A}(Y=0|a^i_j)U_{A}(Y=0, a^i_j).
\end{equation*}
Then, the random optimal action $\delta(\textbf{c})$ selected by the attacker given $\textbf{c}$ is the 
  action maximizing the (random) expected utility
$  \arg\max_{x\in \mathcal{A}}(\Psi_A(x))$.
 We proceed by Monte Carlo (MC) to obtain the required probabilities. At each of $V$ 
  MC iterations $\textcolor{black}{v}$, we sample the random utilities and probabilities and calculate
$ \delta_{\textcolor{black}{v}}(\textbf{c}) = \arg\max_{x\in \mathcal{A}}(\Psi_A^{\textcolor{black}{v}}(x))$. 
 Using $\{\delta_{\textcolor{black}{v}}(\textbf{c})\}^{\textcolor{black}{V}}_{\textcolor{black}{v}=1}$, we estimate the predictive probability $\tau^{1}_j(\textbf{c})$  that the attacker targets our system through any entry point perpetrating attack $j$, given 
  $\textbf{c}$, through

\begin{equation*}
\tau^1_j(\textbf{c}) = P_D(A = a^1_{j,.}|\textbf{c})= \frac{\big|\{\delta_{\textcolor{black}{v}}(\textbf{c})= a^1_j\}\big|}{\textcolor{black}{V}}
\end{equation*}

\noindent and, similarly, for the probability $\tau^{i}(c)$ of the attacker targeting any other system. Such values are compiled in $\textcolor{black}{\tau(\textbf{c}) = ({\tau_1^1(\textbf{c})},\ldots, {\tau_J^1(\textbf{c})},}$ \textcolor{black}{$\tau^i(\textbf{c}))$}. \footnote{This vector serves as input to sample from a multinomial distribution to select whether the attacker targets our system, and if so, which entry block is attacked.}

Additionally, for each attack $j$  we compute

\begin{equation*}
\Gamma^1_{j}(\textbf{c}) =\big(\gamma_{j,1}^1,\ldots, \gamma_{j,K}^1\big) =  \Big( \Big|\{\delta_{m'} = 
a^{1}_{j,1}\}^{\textcolor{black}{V}}_{\textcolor{black}{v}=1}\Big| +\textcolor{black}{1},\ldots, \Big|\{\delta_{\textcolor{black}{v}} = 
a^{1}_{j,K}\}^{\textcolor{black}{V}}_{\textcolor{black}{v}=1}\Big|+\textcolor{black}{1}\Big),
\end{equation*}

\noindent denoting the MC simulations in which the attacker targets each entry with attack \textcolor{black}{$j$}. $\Gamma^{1}_{j}(\textbf{c})$ defines the parameter vector for the Dirichlet distribution to compute entry probabilities  for each attack $j$ given portfolio \textbf{c} in step \ref{alg1:step2} of Algorithm \ref{alg:bis2}, determining the distribution of entry probabilities $P(\textbf{c})$.\footnote{This implements Laplace smoothing \citep{manning2009introduction}, by adding 1 to each component of the vector, to address the issue of null components when sampling from the Dirichlet distribution.}

Algorithm \ref{alg:targeted_attack} expands upon Algorithm \ref{alg:SGAF} by incorporating the previous ARA based method to simulate the impacts of targeted attacks in the system under analysis. 

\begin{algorithm}[H]
\caption{\textcolor{black}{Targeted attack simulation in a general facility}} 
\label{alg:targeted_attack}
\hspace*{\algorithmicindent} \textbf{\textcolor{black}{Input:}} \textcolor{black}{$G$, $\Lambda$, distributions for $Q$,$Q^{s}$}
\hspace*{\algorithmicindent} 
\begin{algorithmic}[1]
\State \textcolor{black}{Estimate attacker behaviour ($\tau$) and target block ($\Gamma_{j}^{1}$)   parameter distributions}
\For{\textcolor{black}{$j \in \{1, \ldots, M\}$}}
\State \textcolor{black}{Generate number $N$ of attacks using $\Lambda$}
\For{\textcolor{black}{$i \in \{1, \ldots, N\}$}}
\State \textcolor{black}{Choose target using $\tau$} \label{step_5}
\If{\textcolor{black}{our system selected}}
\State \textcolor{black}{Generate $\Gamma_{j}^{1}$, $Q$, $Q^{s}$}
\State \textcolor{black}{Simulate attack transit using $\Gamma_{j}^{1}$, $Q$, $Q^s$} \label{step_7}\Comment{\textcolor{black}{Algorithm \ref{alg:bis2}}}
\State \textcolor{black}{Simulate impact $l_i$ from attack} \Comment{\textcolor{black}{Algorithm \ref{alg:MCBI}}}
\Else
\State \textcolor{black}{Impact $l_i=0$}
\EndIf
\EndFor
\State \textcolor{black}{Compute $l^j = \sum_{i=1}^{N}l_i$}
\EndFor
\State \textcolor{black}{\textbf{Return:}  $\{l^j \}_{j=1}^{M}$}
\end{algorithmic}
\end{algorithm}
\noindent It first obtains the vectors $\tau$ and $\Gamma^1_{j}$,
then used to 
compute the impacts of the attacks on the system. We next simulate how many attacks the perpetrator will execute using the arrival process distribution. Then, for each attack it is
 predicted  whether or not the attacker selects our system. If the attacker chooses our system, 
  $\Gamma^1_j$ is used to select which block will be targeted within the system. Subsequently, the impact on the system is computed using Algorithms \ref{alg:bis2} and \ref{alg:MCBI}.

\section{CASE STUDY}
\label{sec:example}

This section presents an illustration of the proposed methodology.
It is a simplification of an actual case but complex enough to reflect  
the required modeling steps. It concerns 
an ADS fleet owner who uses them for rental 
purposes and wishes to improve 
their cybersecurity. 
The key role of AI components in ADS architectures is described 
 in \cite{caballero2021decision}. \cite{baylonreport_1} provides a detailed analysis of cybersecurity issues in relation to ADS. Details on the 
 relevant parameters in this risk analysis over a one-year 
  horizon planning may be found in the Supplementary Materials (SM).

\subsection{Problem description}

Let us briefly describe the relevant elements of the problem. We first characterize the ADS architecture through blocks and levels and 
specify the relevant impacts. Additionally, we identify potential threats, distinguishing between non-adversarial and adversarial ones, recognizing 
their potential perpetrators. Finally, potential defenses not already implemented 
  are presented and we proceed to risk assessment and management.

\begin{figure}[!h]
\centering
\includegraphics[width=.75\linewidth]{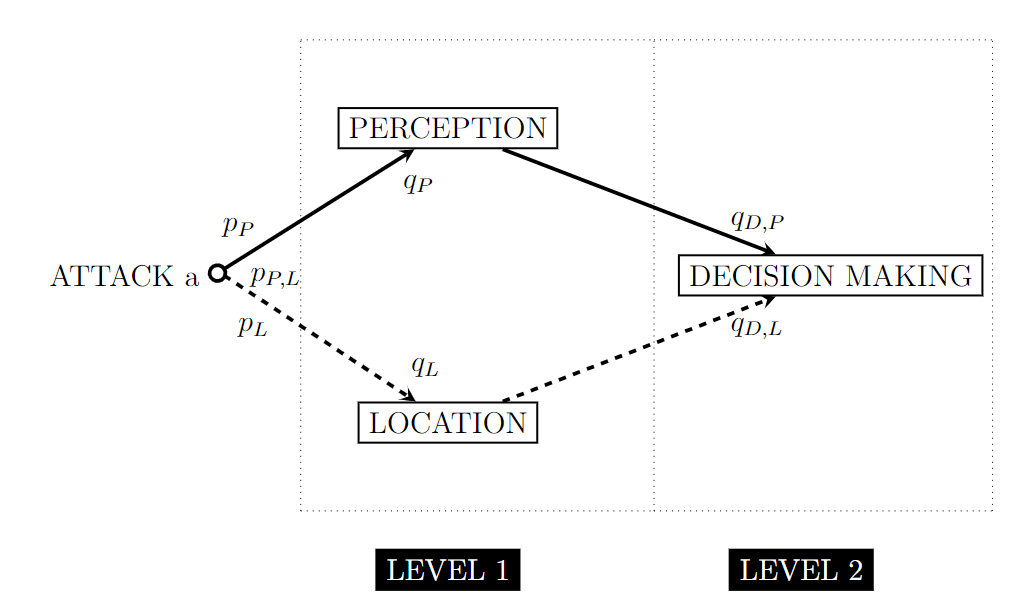}
  \caption{ ADS architecture. Three blocks and two levels.}
  \label{fig:sub1aa}
\end{figure}

\paragraph{ADS architecture.} 
The simplified architecture  (Figure \ref{fig:sub1aa}) includes two level-one blocks susceptible of external attacks. 

\begin{itemize}
    \item \textit{Perception system}, covering sensors (RADAR, LIDAR, Vehicle-to-Vehicle (V2V),...) and processing subsystems  allowing the ADS to determine its relative location with respect to other cars, pedestrians, and obstacles. 
    
    \item \textit{Location system}, covering the systems (GPS,...)
    that estimate the vehicle global position during operations.
\end{itemize}
 Its architecture also includes a level-two block which cannot be attacked directly
\begin{itemize}
    \item \textit{Decision/control system}. Contains the AI applications processing data from  previous blocks to predict the 
    behavior of nearby obstacles and decide appropriate speed 
    and direction to fulfill ADS needs. 
\end{itemize}

\paragraph{Threats.}
We consider the following non-targeted and targeted threats. 

\noindent \subparagraph{Non-targeted threats.}  
  There is a large set of these threats against ADS \citep{taslimasa2023security}. We consider only two of them. 
 \begin{itemize}
     \item 
 {\em {Denial of Service (DoS)}.} Traffic infrastructure may be infected to gain access
 to vehicles through interaction, to prevent users from entering their vehicles by intervening in their door-locking system
 demanding a ransom to regain access. Besides, attackers may also obtain sensitive data from the location block.
   \item {\em Software/hardware supply chain threats (SCTs).} Rogue updates 
   could make an ADS non-operational. Importantly, such updates might not be assured beyond a deadline: obsolescence may pose a threat in certain ADS scenarios \citep{freitas}.
 \end{itemize}

\subparagraph{Targeted threats.}

  We consider the following actors as potential attackers:
   \begin{itemize}

\item \textit{Cyberterrorist group (Cy)}. They could modify traffic signals or interfere with V2V and V2I 
communications to achieve political
notoriety. 

\item \textit{Criminal gang (Cr)}. They carry out their attacks to obtain economic gains through stolen sensitive information.
\end{itemize}

\noindent These groups may target the ADS  through different attacks.  We only consider: 

\begin{itemize}
\item \textit{Adversarial attacks against ML algorithms (AML\_at)}. A perpetrator decides to alter the data plane of vehicular communications or environmental elements to fool the ADS ML algorithms.
For instance, they might modify traffic signals \citep{wei2022adversarial} to 
 trick the perception block and maliciously perturb vehicle behavior. 

\item \textit{Wireless jamming on the control plane (wir\_jam).} A malicious actor could send signals to interfere with the proper functioning of
devices such as the GPS system, 
potentially causing accidents and damage to vehicles and users.\footnote{An interface 
posing a threat to the vehicle is the charging infrastructure for electric cars
\citep{kohler2022brokenwire}.}

\end{itemize}

\noindent Table \ref{tab:att_attacker_tab} summarizes the attacks that actors are  capable to carry out.  We assume that attackers can execute only one type of attack at any given moment.\footnote{This assumption is only relevant for the cyberterrorist group since, unlike the criminal gang, it has the capability to execute both types of attacks.}

\begin{table}[!htbp]
\centering
\begin{tabular}{ccc}
\hline
Attacks/Attacker   & Cyberterrorist  group & Criminal gang \\ \hline
AML\_at & X                     & X             \\
wir\_jam   & X                     &               \\ \hline
\end{tabular}
\caption{Attacks available to malicious actors.}
\label{tab:att_attacker_tab}
\end{table}

\noindent Besides,  two other companies in the market offer similar services. 

\paragraph{Impacts} 
Following Section \ref{sec:impacts},  the owner considers 
 three relevant impacts in this case.

\begin{itemize}

\item \textit{Financial}. They encompass all costs related to loss of sensitive information, including that relevant to ADS users or the rental company. A related current and increasing risk derives from the complexity of the legal ecosystem associated with
ADS liability.\footnote{See the outcomes from UNECE W29: \url{https://unece.org/wp29-introduction}}
Assessed in thousands of euros.

\item \textit{Equipment damage}. This refers to harm inflicted on any ADS component. {Taking into account the importance of software in these vehicles, it would be necessary to consider all threats derived from firmware \citep{halder2020secure} and software management (see \href{https://www.iso.org/standard/81805.html}{ISO/IEC AWI 5888},  \href{https://www.iso.org/standard/77796.html}{ISO 24089:2023}). Assessed 
in thousands of euros}. 

\item \textit{Downtime}, the time the ADS is unavailable due to an attack,
measured in hours.

\end{itemize}

\noindent Table \ref{tab:att_imp_tab} displays the impacts induced by 
different types of attacks.

\begin{table}[!htbp]
\centering
\begin{tabular}{ccccc}
\hline
Attacks/Impacts & Financial & Equipment damage & Downtime  \\ \hline
AML\_at (Cy) & X         & X                & X                 \\
AML\_at (Cr) & X         & X                &                \\
Wireless jam.   & X         & X                & X           \\
DoS             & X         &                  & X          \\
SCTs            & X         & X                & X             \\ \hline
\end{tabular}
\caption{Impacts deemed relevant for targeted and untargeted threats.}
\label{tab:att_imp_tab}
\end{table}

\paragraph{Defenses}
The defenses considered for
risk management purposes not yet implemented in the ADS 
are: a firewall and internet gateway \textit{(FwGw)}; a robust AML module \textit{(AML)}; and, a patch management, IDS and vulnerability scanner \textit{(PmVs)}.  Table
\ref{tab:att_def_tab} displays the defences that are effective against 
various types of attacks.

\begin{table}[H]
\centering
\begin{tabular}{cccc}
\hline
Attacks/Defenses   & FwGw & AML  & PmVs \\ \hline
Adversarial attack &      & X    &      \\
Wireless jamming   & X    &      & X    \\
DoS                & X    &      & X    \\
SCTs               & X    & X    & X    \\ \hline
\end{tabular}
\caption{Defense effectiveness and costs.}
\label{tab:att_def_tab}
\end{table}

\noindent Cyber insurance options are also
available.  Two products are considered: a basic one ($A$) covering equipment damage occurring to the vehicle and an advanced one ($B$) that, additionally, covers costs related to downtime. Table \ref{tab:prod_att}
displays which impacts are mitigated by each cyber insurance product.

\begin{table}[!htbp]
\centering
\begin{tabular}{ccccc}
\hline
  & Finan. & \begin{tabular}[c]{@{}c@{}}Equip.\\ damage\end{tabular} & Downt. \\ \hline
A &        & X                                                       &               \\
B &        &  X                                                       & X             \\ \hline
\end{tabular}

\caption{Impacts mitigated by each cyber insurance product and its cost.}
\label{tab:prod_att}
\end{table}

\paragraph{Constraints}
The maximum protection budget is 3400 euros. Legislation 
requires at least a type $A$ cyber insurance product to be included at the minimum. 

\subsection{Risk analysis}
\label{sec:risk_assesment}
 This section focuses on evaluating the risk associated with the system within its initial configuration (absence of new protections and insurance product $A$). We apply the framework detailed in Sections \ref{sec:attack_simulation} and \ref{sec:targeted_att}, utilizing the parametric setup outlined in the SM. The objective is to identify the portfolio that displays the highest efficacy in mitigating the risks within the system.

\begin{figure}[h!]
\includegraphics[width=0.6\textwidth]{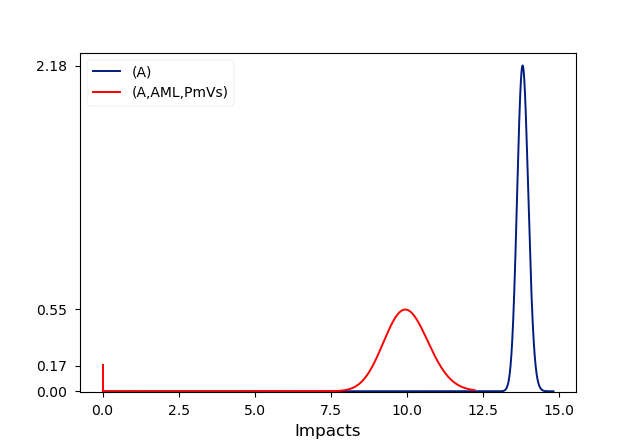}
\centering
\caption{ADS risk assessment loss curves on a logarithmic scale for initial (blue) and optimal (red) portfolio configurations.}
\label{kkjlo}
\end{figure}

\paragraph{Risk assessment.} Figure \ref{kkjlo} (\textcolor{black}{dark blue line}) illustrates the resulting loss curve in euros (logarithmic scale) with the initial configuration, obtained with an MC sample size of 10,000. The probability of experiencing no loss is zero, and positive losses 
 are approximated through a gamma distribution. From this distribution, we obtain a 95\% VaR at 1.32 and a 95\% CVaR at 1.498 million euros. Given that the risk is considered too high, we proceed on to determine the best security portfolio through a risk management stage.

\paragraph{Risk management.} Consider now the RM problem with parameters as in the SM. 
Only 12 portfolios out of 16  are feasible. As this number is small, it is reasonable to undertake
the evaluation of all 12 of them. \footnote{It takes about 10 minutes in a standard laptop.}
Their expected utilities are assessed by MC  
 as in Algorithm \ref{alg:generate_sample}.
  In principle, we use $\rho =  10^{-7}$ as risk aversion coefficient and, again,
an MC sample size of 10000.  Table \ref{kkxmas} presents the three best 
portfolios together with their expected loss, cost, and utility. 
The optimal portfolio consists of adopting Insurance A and installing the AML and PmVs modules;
 the second best is (B, FwGw, AML), whereas (A, FwGw, AML) is
 the third preferred portfolio. 

\begin{table}[h!]
\centering
\begin{tabular}{cccc}
\hline
Portfolio        & Expected loss & Cost & Expected Utility \\ \hline
$ A, AML, PmVs $ & 22834.59  & 2300 & -0.0025 \\
$ B, FwGw, AML$ & 43918.99  & 2950 & -0.0047 \\
$ A, FwGw, AML $ & 48185.75  & 1800 & -0.0050 \\ \hline
\end{tabular}
\caption{Three best portfolios.}
\label{kkxmas}
\end{table}

\noindent Figure \ref{kkjlo} (red) displays the predictive loss curve for the best portfolio.
  Observe that the probability of zero loss with the optimal portfolio is 0.174. 
 The positive part is fit with one gamma component. The 95\% Var and 95\% CVaR would be 59520 and 72756 euros, respectively. Observe, therefore, the significant risk reduction 
 attained when implementing the selected portfolio, achieving a high level of protection given the budget available.

 We performed extensive sensitivity analysis to assess the robustness of the output
 to various parameters. In particular, we considered sensitivity to the risk 
 aversion coefficient $\rho $ by varying it 
 within a grid in the range $[10^{-7}, 10^{-3}]$.
 The same optimal portfolio is preserved with only a switch between the second
 and third portfolios, as $\rho$ gets bigger than $10^{-4}$, suggesting robustness of the response.

\section{DISCUSSION AND OPEN ISSUES}

Motivated by the EU AI Act, the NIST AIRMF and the CRFM report, 
we have described new cybersecurity risk analysis issues that emerge in systems with AI components. In particular, we described the challenges that this technology brings in regarding assets, impacts, controls, and targeted attacks and provided a broad framework for 
their risk analysis. Under the proposed approach, we structure a system through blocks in different levels, with links indicating potential attack transit flows 
through information or transaction exchanges.
We also introduced a scheme to simulate an attack transit within the system, allowing us to obtain risk indicators and modeled the problem in which risk mitigations to be added to a security portfolio have to be selected, including cyber insurance, to minimize system risks, as illustrated with a case referring to ADSs. 

 Several measures have been developed recently to treat the safety, ethical, bias, privacy and fairness threats that are becoming of major concern in the AI domain, including, for instance, mechanisms to enforce anti-discrimination policies, privacy controls or the monitoring of sensitive content to ensure safety. Although these are not exactly cybersecurity controls, they can be actually integrated into our framework. The decision-support process regarding their selection
 would be similar, aggregating them as procedural, technical, or physical controls against unwarranted actions. This can be used in a broader risk analysis process that includes not only cybersecurity, but also safe and fair uses of the AI system of interest. 

Beyond as a risk analysis framework,
we may use our risk assessment proposal to determine where at the EU AI Act four tier risk ladder a
system is. In particular, we could study how vulnerable 
 a system is and map its derived risks to one of the tiers, understanding whether it is only advisable to implement additional security controls. The proposed methodology is flexible enough to cover the impacts, attacks, and defenses for AI components besides constraints over the mitigations in relation to compliance with laws  or standards. This flexibility contributes to defining a robust framework to integrate advanced risk analysis and protection methods for AI applications, but also to support certification programs as demanded in the EU Cybersecurity and Cyber-resilience Acts \citep{cihon2019standards}. It may be used as well in a proactive manner
 by checking the impact of protection measures over the assessed risk through a what-if type of analysis.

As our case study shows, admittedly our proposed framework is rather technical and demands intensive modelling work.
 However, the values at stake are 
so important that the extra effort should be definitely worth it  
when compared to the simplistic approaches emerging in the field as replicas
of major cybersecurity risk 
analysis standards. To facilitate its implementation 
a system could be developed, possibly adopting the ENISA terminology set up in \cite{ENISA2023},
extended with the novel AI ingredients reflected in this document.
Importantly, most of these ingredients will be common across many systems and their corresponding distributions will be similar for blocks with the same structure and configuration, thus alleviating elicitation tasks and allowing the development of templates of blocks and distributions.

 An additional aspect of AI product protection is the inclusion of security as well as other risk-related objectives (e.g., privacy, safety, fairness) in their design and development. Our framework could be used to address the selection of secure or other risk-related features in the design of AI based systems, thus embedding 
 a security-by-design approach matching the efforts in securisation and riskification \citep{backman2023risk} in the context of AI and cybersecurity.

 \section*{ACKNOWLEDGEMENTS}

Blanked.

\bibliography{main}

\begin{thebibliography}{}

\bibitem[Agarwala et~al., 2020]{agarwala2020supervisory}
Agarwala, G., Latorre, A., Raffel, S., Mehta, R., Zhao, J., Nurullayev, A., Clark, B., \& Tang, R. (2020).
\newblock \textit{Supervisory expectations and sound model risk management practices for artificial intelligence and machine learning}.
\newblock {\em {\normalfont Ernst {\&} Young}}.
\newblock \url{https://assets.ey.com/content/dam/ey-sites/ey-com/en_us/topics/banking-and-capital-markets/ey-mrm-ai-ml.pdf}\killpunct.

\bibitem[Backman, 2023]{backman2023risk}
Backman, S. (2023).
\newblock Risk vs. threat-based cybersecurity: The case of the {EU}.
\newblock {\em European Security}, 32(1), \killpunct:85--103.
\newblock \url{https://doi.org/10.1080/09662839.2022.2069464} \killpunct.

\bibitem[Banks et~al., 2022]{banks2022adversarial}
Banks, D., Gallego, V., Naveiro, R., \& R{\'i}os~Insua, D. (2022).
\newblock Adversarial risk analysis: An overview.
\newblock {\em Wiley Interdisciplinary Reviews: Computational Statistics}, 14(1), \killpunct:Article e1530.
\newblock \url{https://doi.org/10.1002/wics.1530} \killpunct.

\bibitem[Baylon, 2017]{baylonreport_1}
Baylon, C. (2017).
\newblock \textit{Connected Cars: Opportunities and Risk for the Insurance Company}.
\newblock {\em {\normalfont AXA}}.

\bibitem[Biggio \& Roli, 2018]{BIGGIO2018317}
Biggio, B. \& Roli, F. (2018).
\newblock Wild patterns: Ten years after the rise of adversarial machine learning.
\newblock {\em Pattern Recognition}, 84\killpunct:, 317--331.
\newblock \url{https://doi.org/10.1016/j.patcog.2018.07.023} \killpunct.

\bibitem[Boloor et~al., 2019]{boloor2019simple}
Boloor, A., He, X., Gill, C., Vorobeychik, Y., \& Zhang, X. (2019).
\newblock Simple physical adversarial examples against end-to-end autonomous driving models.
\newblock {\em Proceedings of the 2019 {IEEE} International Conference on Embedded Software and Systems (ICESS)}, \killpunct0\killpunct:1--7.
\newblock \url{https://doi.org/10.1109/ICESS.2019.8782514} \killpunct.

\bibitem[Caballero et~al., 2023]{caballero2021decision}
Caballero, W.~N., R{\'i}os~Insua, D., \& Banks, D. (2023).
\newblock Decision support issues in automated driving systems.
\newblock {\em International Transactions in Operational Research}, 30(3), \killpunct:1216--1244.
\newblock \url{htpps://doi.org/10.1111/itor.12936}\killpunct.

\bibitem[Cameron, 2010]{cameron2010vsl}
Cameron, T.~A. (2010).
\newblock Euthanizing the value of a statistical life.
\newblock {\em Review of Environmental Economics and Policy}, 4(2), \killpunct:161--178.
\newblock \url{https://doi.org/10.1093/reep/req010}\killpunct.

\bibitem[Cihon, 2019]{cihon2019standards}
Cihon, P. (2019).
\newblock \textit{Standards for {AI} governance: International standards to enable global coordination in {AI} research \& development}.
\newblock {\em {\normalfont Future of Humanity Institute, University of Oxford}}.
\newblock \url{https://www.fhi.ox.ac.uk/wp-content/uploads/Standards_-FHI-Technical-Report.pdf}\killpunct.

\bibitem[Comiter, 2019]{COMITER2019}
Comiter, M. (2019).
\newblock \textit{Attacking artificial intelligence: {AI}’s security vulnerability and what policymakers can do about it}.
\newblock {\em {\normalfont {Belfer Center for Science and International Affairs, Harvard Kennedy School}}}.
\newblock \url{https://www.belfercenter.org/sites/default/files/2019-08/AttackingAI/AttackingAI.pdf}\killpunct.

\bibitem[Couce-Vieira et~al., 2020]{couce2020assessing}
Couce-Vieira, A., Insua, D.~R., \& Kosgodagan, A. (2020).
\newblock Assessing and forecasting cybersecurity impacts.
\newblock {\em Decision Analysis}, 17(4), \killpunct:356--374.
\newblock \url{https://doi.org/10.1287/deca.2020.0418} \killpunct.

\bibitem[Cox~Jr, 2008]{anthony2008s}
Cox~Jr, L.~A. (2008).
\newblock What's wrong with risk matrices?
\newblock {\em Risk Analysis}, 28(2), \killpunct:497--512.
\newblock \url{https://doi.org/10.1111/j.1539-6924.2008.01030.x}\killpunct.

\bibitem[De~Freitas et~al., 2021]{freitas}
De~Freitas, J., Censi, A., Smith, B.~W., Di~Lillo, L., Anthony, S.~E., \& Frazzoli, E. (2021).
\newblock From driverless dilemmas to more practical commonsense tests for automated vehicles.
\newblock {\em Proceedings of the National Academy of Sciences}, 118(11), \killpunct:Article e2010202118.
\newblock \url{https://doi.org/10.1073/pnas.2010202118}\killpunct.

\bibitem[ETSI, 2022]{ETSI}
ETSI (2022).
\newblock \textit{ETSI GR SAI 006 v1.1.1: Securing Artificial Intelligence (SAI): The role of hardware in security of AI.}
\newblock \url{https://www.etsi.org/deliver/etsi_gr/SAI/001_099/006/01.01.01_60/gr_SAI006v010101p.pdf}\killpunct.

\bibitem[{European Commission}, 2021]{commissie2021proposal}
{European Commission} (2021).
\newblock \textit{Proposal for a Regulation of the European Parliament and of the Council laying down harmonised rules on Artificial Intelligence (Artificial Intelligence Act) and amending certain Union legislative acts. COM/2021/206 final [Document 52021PC0206].}
\newblock \url{https://eur-lex.europa.eu/legal-content/EN/TXT/?uri=celex%3A52021PC0206}\killpunct.

\bibitem[{European Parliament} et~al., 2012]{commissie2012charter}
{European Parliament}, {European Council}, \& {European Commission} (2012).
\newblock \textit{Charter of Fundamental Rights of the European Union. OJ C 326, 26.10.2012 [Document C2012/326/02]}.
\newblock \url{https://eur-lex.europa.eu/LexUriServ/LexUriServ.do?uri=OJ:C:2012:326:0391:0407:EN:PDF}\killpunct.

\bibitem[{Federal Bureau of Investigation}, 2022]{crimereport}
{Federal Bureau of Investigation} (2022).
\newblock \textit{Internet crime report 2022. U.S. Department of Justice}.
\newblock \url{https://www.ic3.gov/Media/PDF/AnnualReport/2022_IC3Report.pdf}\killpunct.

\bibitem[Freeman~III et~al., 2014]{freeman2014vsl}
Freeman~III, A.~M., Herriges, J.~A., \& Kling, C.~L. (2014).
\newblock {\em The Measurement of Environmental and Resource Values: Theory and Methods {\normalfont (3rd ed.).}}
\newblock Routledge.
\newblock \url{https://doi.org/10.4324/9781315780917}\killpunct.

\bibitem[Gallego et~al., 2023]{gallego}
Gallego, V., Naveiro, R., Redondo, A., R{\'i}os~Insua, D., \& Ruggeri, F. (2023).
\newblock \textit{Protecting Classifiers From Attacks. {A} {B}ayesian Approach}.
\newblock {\em {\normalfont ArXiv}}.
\newblock \url{https://arxiv.org/abs/2004.08705}\killpunct.

\bibitem[Gallego et~al., 2021]{gallego2021ai}
Gallego, V., Naveiro, R., Roca, C., R{\'i}os~Insua, D., \& Campillo, N.~E. (2021).
\newblock {AI} in drug development: A multidisciplinary perspective.
\newblock {\em Molecular Diversity}, 25(3), \killpunct:1461--1479.
\newblock \url{https://doi.org/10.1007/s11030-021-10266-8}\killpunct.

\bibitem[Gallego \& R{\'i}os~Insua, 2022]{gallego2022current}
Gallego, V. \& R{\'i}os~Insua, D. (2022).
\newblock Current advances in neural networks.
\newblock {\em Annual Review of Statistics and Its Application}, 9, \killpunct:197--222.
\newblock \url{https://doi.org/10.1146/annurev-statistics-040220-112019}\killpunct.

\bibitem[Goldwasser et~al., 2022]{goldwasser2022planting}
Goldwasser, S., Kim, M.~P., Vaikuntanathan, V., \& Zamir, O. (2022).
\newblock Planting undetectable backdoors in machine learning models.
\newblock {\em Proceedings of the 2022 {IEEE} 63rd Annual Symposium on Foundations of Computer Science ({FOCS})}, 931--942.
\newblock \url{https://doi.org/10.1109/FOCS54457.2022.00092}\killpunct.

\bibitem[Gonz{\'a}lez-Ortega et~al., 2018]{GonzalezOrtega2018}
Gonz{\'a}lez-Ortega, J., Radovic, V., \& R{\'\i}os~Insua, D. (2018).
\newblock Utility elicitation.
\newblock {\em {\normalfont In L.C. Dias, A. Morton, \& J. Quigley (Eds.),} Elicitation: The Science and Art of Structuring Judgement {\normalfont (pp. 241-264)}. {\normalfont Springer}}.
\newblock \url{https://doi.org/10.1007/978-3-319-65052-4_10}\killpunct.

\bibitem[Halder et~al., 2020]{halder2020secure}
Halder, S., Ghosal, A., \& Conti, M. (2020).
\newblock Secure over-the-air software updates in connected vehicles: A survey.
\newblock {\em Computer Networks}, 178, Article 107343.
\newblock \url{https://doi.org/10.1016/j.comnet.2020.107343}\killpunct.

\bibitem[Ham, 2021]{ham2021toward}
Ham, J. V.~D. (2021).
\newblock Toward a better understanding of “cybersecurity”.
\newblock {\em Digital Threats: Research and Practice}, 2(3), 1--3.
\newblock \url{https://doi.org/10.1145/3442445} \killpunct.

\bibitem[Hanea et~al., 2021]{hanea2021expert}
Hanea, A.~M., Nane, G.~F., Bedford, T., \& French, S. (Eds.). (2021).
\newblock {\em Expert Judgement in Risk and Decision Analysis}.
\newblock Springer Cham.
\newblock \url{https://doi.org/10.1007/978-3-030-46474-5}\killpunct.

\bibitem[Hasan \& Salah, 2019]{hasan2019combating}
Hasan, H.~R. \& Salah, K. (2019).
\newblock Combating deepfake videos using blockchain and smart contracts.
\newblock {\em {IEEE} Access}, 7, \killpunct:41596--41606.
\newblock \url{https://doi.org/10.1109/ACCESS.2019.2905689} \killpunct.

\bibitem[Helberger \& Diakopoulos, 2023]{helberger2023chatgpt}
Helberger, N. \& Diakopoulos, N. (2023).
\newblock {ChatGPT} and the {AI} act.
\newblock {\em Internet Policy Review}, 12(1).
\newblock \url{https://doi.org/10.14763/2023.1.1682}\killpunct.

\bibitem[{Information Security Forum}, 2016]{isf2003}
{Information Security Forum} (2016).
\newblock \textit{Information Risk Assessment Methodology 2 (IRAM 2)}.
\newblock \url{https://www.securityforum.org/solutions-and-insights/information-risk-assessment-methodology-2-iram2/}\killpunct.

\bibitem[{International Organization for Standardization}, 2020]{iso24028}
{International Organization for Standardization} (2020).
\newblock {\textit{Information technology -- Artificial intelligence -- Overview of trustworthiness in artificial intelligence} (ISO Standard No. TR 24028:2020)}.
\newblock \url{https://www.iso.org/standard/77608.html}\killpunct.

\bibitem[{International Organization for Standardization}, 2022]{iso5723}
{International Organization for Standardization} (2022).
\newblock {{\textit{Trustworthiness -- Vocabulary}} (ISO Standard No. TS 5723:2022)}.
\newblock \url{https://www.iso.org/standard/81608.html}\killpunct.

\bibitem[Kaloudi \& Li, 2020]{kaloudi2020ai}
Kaloudi, N. \& Li, J. (2020).
\newblock The {AI}-based cyber threat landscape: A survey.
\newblock {\em ACM Computing Surveys}, 53 (1), \killpunct:1--34.
\newblock \url{https://doi.org/10.1145/3372823} \killpunct.

\bibitem[Keeney \& Gregory, 2005]{keeney2005selecting}
Keeney, R.~L. \& Gregory, R.~S. (2005).
\newblock Selecting attributes to measure the achievement of objectives.
\newblock {\em Operations Research}, 53(1), \killpunct:1--11.
\newblock \url{https://doi.org/10.1287/opre.1040.0158}\killpunct.

\bibitem[K{\"o}hler et~al., 2023]{kohler2022brokenwire}
K{\"o}hler, S., Baker, R., Strohmeier, M., \& Martinovic, I. (2023).
\newblock Brokenwire: Wireless disruption of {CCS} electric vehicle charging.
\newblock {\em Network and Distributed System Security ({NDSS}) Symposium 2023}.
\newblock \url{https://doi.org/10.14722/ndss.2023.23251}\killpunct.

\bibitem[Leone, 2023]{leone2023spiral}
Leone, M. (2023).
\newblock The spiral of digital falsehood in deepfakes.
\newblock {\em International Journal for the Semiotics of Law -- Revue internationale de S{\'e}miotique juridique}, 36(2), \killpunct:385--405.
\newblock \url{https://doi.org/10.1007/s11196-023-09970-5}\killpunct.

\bibitem[Madiega, 2021]{madiega2023artificial}
Madiega, T. (2021).
\newblock \textit{Artificial intelligence act. Briefing of EU Legislation in Progress, Document PE 698.792.}
\newblock {\em {\normalfont European Parliamentary Research Service}}.
\newblock \url{https://www.europarl.europa.eu/RegData/etudes/BRIE/2021/698792/EPRS_BRI(2021)698792_EN.pdf}\killpunct.

\bibitem[Manning et~al., 2008]{manning2009introduction}
Manning, C.~D., Raghavan, P., \& Sch{\"u}tze, H. (2008).
\newblock {\em An Introduction to Information Retrieval}.
\newblock Cambridge University Press.
\newblock \url{https://nlp.stanford.edu/IR-book/html/htmledition/irbook.html}\killpunct.

\bibitem[Masure et~al., 2020]{masure2020comprehensive}
Masure, L., Dumas, C., \& Prouff, E. (2020).
\newblock A comprehensive study of deep learning for side-channel analysis.
\newblock {\em {IACR} Transactions on Cryptographic Hardware and Embedded Systems}, 2020(1), \killpunct:348--375.
\newblock \url{https://doi.org/10.13154/tches.v2020.i1.348-375}\killpunct.

\bibitem[McDaniel, 2022]{mcdaniel2022sustainability}
McDaniel, P.~D. (2022).
\newblock Sustainability is a security problem.
\newblock {\em Proceedings of the 28th {ACM SIGSAC} Conference on Computer and Communications Security (CCS'22)}, 9--10.
\newblock \url{https://doi.org/10.1145/3548606.3559396} \killpunct.

\bibitem[Naveiro et~al., 2019]{naveiro2019adversarial}
Naveiro, R., Redondo, A., R{\'\i}os~Insua, D., \& Ruggeri, F. (2019).
\newblock Adversarial classification: An adversarial risk analysis approach.
\newblock {\em International Journal of Approximate Reasoning}, 113, \killpunct:133--148.
\newblock \url{https://doi.org/10.1016/j.ijar.2019.07.003}\killpunct.

\bibitem[{NIST}, 2023]{nist2022ai}
{NIST} (2023).
\newblock \textit{NIST AI 100-1: Artificial Intelligence Risk Management Framework (AI RMF 1.0)}. {U.S. Department of Commerce.}
\newblock \url{https://doi.org/10.6028/NIST.AI.100-1}\killpunct.

\bibitem[Oliynyk et~al., 2023]{oliynyk2023know}
Oliynyk, D., Mayer, R., \& Rauber, A. (2023).
\newblock I know what you trained last summer: A survey on stealing machine learning models and defences.
\newblock {\em {ACM} Computing Surveys}, 55(14s), \killpunct:1--41.
\newblock \url{https://doi.org/10.1145/3595292} \killpunct.

\bibitem[{Organisation for Economic Co-operation and Development}, 2019]{oecd2019ai}
{Organisation for Economic Co-operation and Development} (2019).
\newblock \textit{Recommendation of the Council on Artificial Intelligence. Document OECD/LEGAL/0449}.
\newblock \url{https://legalinstruments.oecd.org/en/instruments/oecd-legal-0449}\killpunct.

\bibitem[Papadatos et~al., 2023]{ENISA2023}
Papadatos, K., Rantos, K., Markrygergeou, A., Koulouris, K., Klontza, S., Lamabrinoudakis, C., Grirzalis, S., Xenakis, C., Katsikas, S., Karyda, M., Tsochou, A., \& Zacharis, A. (2023).
\newblock \textit{Interoperable {EU} Risk Management Toolbox.} {European Union Agency for Cybersecurity ({ENISA})}.
\newblock \url{https://doi.org/10.2824/68948}\killpunct.

\bibitem[Papoulis \& Unnikrishna~Pillai, 2002]{papoulis2002probability}
Papoulis, A. \& Unnikrishna~Pillai, S. (2002).
\newblock {\em Probability, Random Variables and Stochastic Processes (4th ed.)}.
\newblock McGraw-Hill Europe.

\bibitem[Powell, 2019]{powell2019unified}
Powell, W.~B. (2019).
\newblock {\em Reinforcement Learning and Stochastic Optimization}.
\newblock John Wiley \& Sons.

\bibitem[Redondo \& R{\'\i}os~Insua, 2020]{redondo2020protecting}
Redondo, A. \& R{\'\i}os~Insua, D. (2020).
\newblock Protecting from malware obfuscation attacks through adversarial risk analysis.
\newblock {\em Risk Analysis}, 40(12), \killpunct:2598--2609.
\newblock \url{https://doi.org/10.1111/risa.13567}\killpunct.

\bibitem[R{\'i}os \& R{\'i}os~Insua, 2012]{rios2012adversarial}
R{\'i}os, J. \& R{\'i}os~Insua, D. (2012).
\newblock Adversarial risk analysis for counterterrorism modeling.
\newblock {\em Risk Analysis}, 32(5), \killpunct:894--915.
\newblock \url{https://10.1111/j.1539-6924.2011.01713.x}\killpunct.

\bibitem[R{\'i}os~Insua et~al., 2021a]{insua2020security}
R{\'i}os~Insua, D., Baylon, C., \& Vila, J. (Eds.). (2021a).
\newblock {\em Security Risk Models for Cyber Insurance}.
\newblock CRC Press.

\bibitem[R{\'i}os~Insua et~al., 2021b]{rios2021adversarial}
R{\'i}os~Insua, D., Couce-Vieira, A., Rubio, J.~A., Pieters, W., Labunets, K., \& G.~Rasines, D. (2021b).
\newblock An adversarial risk analysis framework for cybersecurity.
\newblock {\em Risk Analysis}, 41(1), \killpunct:16--36.
\newblock \url{https://doi.org/10.1111/risa.13331}\killpunct.

\bibitem[R{\'i}os~Insua et~al., 2023]{advclas}
R{\'i}os~Insua, D., Naveiro, R., Gallego, V., \& Poulos, J. (2023).
\newblock Adversarial machine learning: {B}ayesian perspectives.
\newblock {\em Journal of the American Statistical Association}, 118(543), \killpunct:2195--2206.
\newblock \url{https://doi.org/10.1080/01621459.2023.2183129}\killpunct.

\bibitem[Sanyal et~al., 2022]{sanyal2022towards}
Sanyal, S., Addepalli, S., \& Babu, R.~V. (2022).
\newblock Towards data-free model stealing in a hard label setting.
\newblock {\em Proceedings of the 2022 {IEEE/CVF} Conference on Computer Vision and Pattern Recognition (CVPR)}, 15263--15272.
\newblock \url{https://doi.org/10.1109/CVPR52688.2022.01485}\killpunct.

\bibitem[{Stanford Center for Research on Foundation Models}, 2021]{Stanford2021FoundationModels}
{Stanford Center for Research on Foundation Models} (2021).
\newblock \textit{On the Opportunities and Risks of {F}oundation {M}odels}, {Stanford University}.
\newblock \url{https://doi.org/10.48550/arXiv.2108.07258}\killpunct.

\bibitem[Szegedy et~al., 2014]{szegedy2013intriguin}
Szegedy, C., Zaremba, W., Sutskever, I., Bruna, J., Erhan, D., Goodfellow, I., \& Fergus, R. (2014).
\newblock Intriguing properties of neural networks.
\newblock {\em Proceedings of the 2nd International Conference on Learning Representations ({ICLR 2014})}.

\bibitem[Taslimasa et~al., 2023]{taslimasa2023security}
Taslimasa, H., Dadkhah, S., Neto, E. C.~P., Xiong, P., Ray, S., \& Ghorbani, A.~A. (2023).
\newblock Security issues in {I}nternet of vehicles ({IoV}): A comprehensive survey.
\newblock {\em Internet of Things}, Article 100809. \killpunct.
\newblock \url{https://doi.org/10.1016/j.iot.2023.100809}\killpunct.

\bibitem[{The White House}, 2023]{whitehouse}
{The White House} (2023).
\newblock \textit{Executive Order on the Safe, Secure, and Trustworthy Development and Use of Artificial Intelligence}.
\newblock \textit{Executive Order 14110 of October 30, 2023}.
\newblock \url{https://www.federalregister.gov/d/2023-24283}\killpunct.

\bibitem[Thekdi \& Aven, 2023]{THEKDI2023106129}
Thekdi, S. \& Aven, T. (2023).
\newblock {Is risk analysis a source of misinformation? The undermining effects of uncertainty on credibility}.
\newblock {\em Safety Science}, 163, \killpunct:Article 106129.
\newblock \url{https://doi.org/10.1016/j.ssci.2023.106129}\killpunct.

\bibitem[Urbina et~al., 2022]{urbina2022dual}
Urbina, F., Lentzos, F., Invernizzi, C., \& Ekins, S. (2022).
\newblock Dual use of artificial-intelligence-powered drug discovery.
\newblock {\em Nature Machine Intelligence}, 4(3), \killpunct:189--191.
\newblock \url{https://doi.org/10.1038/s42256-022-00465-9}\killpunct.

\bibitem[Viscusi, 2020]{viscusi2020pricing}
Viscusi, W.~K. (2020).
\newblock {Pricing the global health risks of the COVID-19 pandemic}.
\newblock {\em Journal of Risk and Uncertainty}, 61(2), \killpunct:101--128.
\newblock \url{https://doi.org/10.1007/s11166-020-09337-2} \killpunct.

\bibitem[Vorobeichyk \& Kantarcioglu, 2019]{vorobeichikantar}
Vorobeichyk, Y. \& Kantarcioglu, M. (2019).
\newblock {\em Adversarial Machine Learning}.
\newblock Morgan \& Claypool.

\bibitem[Wei et~al., 2023]{wei2022adversarial}
Wei, X., Guo, Y., \& Yu, J. (2023).
\newblock Adversarial sticker: A stealthy attack method in the physical world.
\newblock {\em {IEEE} Transactions on Pattern Analysis and Machine Intelligence}, 45(3), \killpunct:2711--2725.
\newblock \url{https://doi.org/10.1109/TPAMI.2022.3176760}\killpunct.

\bibitem[Williams, 1994]{williams1994purdue}
Williams, T.~J. (1994).
\newblock The {P}urdue enterprise reference architecture.
\newblock {\em Computers in Industry}, 24(2-3), \killpunct:141--158.
\newblock \url{https://doi.org/10.1016/0166-3615(94)90017-5}\killpunct.

\bibitem[Wiper et~al., 2001]{wiper}
Wiper, M., R{\'i}os~Insua, D., \& Ruggeri, F. (2001).
\newblock Mixtures of gamma distributions with applications.
\newblock {\em Journal of Computational and Graphical Statistics}, 10(3), \killpunct:440--454.
\newblock \url{https://doi.org/10.1198/106186001317115054}\killpunct.

\end{thebibliography}

\appendix

\pagebreak 

\section*{SUPPLEMENTARY MATERIALS:  \\ PARAMETRIC SETUP}
\label{sec:parametric_setup}

This appendix outlines the models and parameters adopted in the proposed scenario. We first introduce the notation used as well as some general considerations concerning ADS. Then, we indicate the distributions modeling the attacks that might affect the system.

\subsection*{A.1 Notation and general considerations}

\paragraph{Portfolio.} 
To simplify the notation, we do not include the insurance products in $\textbf{c}$. Thus,
portfolios ${\bf c}$ will have the structure ($FwGw$, $AML$, $PmVs$), with  $FwGw$, $AML$, and $PmVs$  being 1 if the corresponding control is implemented and 0, otherwise. Thus, the initial portfolio will be $(0,0,0)$. Additionally, we include only portfolios relevant to handle the corresponding attack. For instance, when referring to DoS attacks, the AML module is not included in the portfolio as it offers no protection against such threat. 
The implementation costs of the controls are, respectively, \textcolor{black}{1250} \hspace{0.01cm}\euro
(FwGw), \textcolor{black}{300} \hspace{0.01cm}\euro  \hspace{0.01cm} (AML) and \textcolor{black}{1750}\euro \hspace{0.01cm} (PmVs).\footnote{Costs and other parameters derived from public
sources or extracted from one expert in cybersecurity and one expert in
financial planning in the team, using standard expert judgement elicitation techniques
\citep{hanea2021expert}. For a few representative parameters, we comment some of their implications.} 

\paragraph{\textit{Access probabilities.}} The probability that an attack uniquely accesses the system through the perception (location) block is denoted $p_{P}$ ($p_{L}$). The access probability through both blocks, $p_{P,L}$.

\paragraph{\textit{Non-protection probabilities}.} $q_{P}$ 
($q_{L}$) refers to the PNP of the perception
(location) system from an external attack.
 $q_{D,P}$ ($q_{D,L}$) designates the PNP of the decision-making block from the perception (location) block.   
   
\paragraph{\textit{Impacts.}} Denote by $l_{FI}$ the financial impact of an attack on the system; $l_{ED}$,
 the equipment damage impact, and $l_{DT}$, the downtime  impact.  
 We assume that the rental car company experiences a cost of \textcolor{black}{100 euros} for each downtime hour of a single ADS. $l(\textbf{c})$ indicates the loss
when portfolio $\textbf{c}$ is implemented.  
 $l_{FI}$ is a \emph{global} impact, whereas $l_{DT}$ and $l_{ED}$ are \emph{separable}: we denote $l_{i_j}$ with $i\in\{ED,DT\}$ and $j\in\{L,P,D\}$ as the $i$-th impact on the $j$-th block. $l_{FI}$, $l_{DT}$, and $l_{ED}$ will be sampled from Gamma distributions.\footnote{For
 continuous non-negative quantities assumed to be unimodal, we use Gamma distributions for flexibility reasons as they  adopts a wide variety of locations and asymmetries contingent on 
  their parametrization \citep{papoulis2002probability} $Gamma(a,p)$
with mean $a\times p$.} Their parameters are detailed in subsequent sections, and we apply aggregation rules from Section \ref{sec:sim_agg_imp}.
  
\paragraph{\textit{Insurance products.}} \textcolor{black}{ Product A reduces the economic impact over equipment damage by $65\%$. If product B is purchased, $70\%$ of the total expenses resulting from equipment damage and downtime will be covered.} Table \ref{tab:ins_portfolio} displays the costs of the insurance products on an ADS depending on adopted portfolio.
\begin{table}[h]
\centering
\textcolor{black}{
\begin{tabular}{ccccccccc}
\cline{2-9}
\hline
\multicolumn{1}{l}{} & (0,0,0) & (1,0,0) & (0,1,0) & (0,0,1) & (1,1,0) & (1,0,1) & (0,1,1) & (1,1,1) \\ \hline
A                    & \textcolor{black}{600}    & \textcolor{black}{500}     & \textcolor{black}{500}     & \textcolor{black}{500}     & \textcolor{black}{250}     & \textcolor{black}{250}     & \textcolor{black}{250}     & \textcolor{black}{150}     \\
B                    & \textcolor{black}{1800}    & \textcolor{black}{1600}    & \textcolor{black}{1600}    & \textcolor{black}{1600} & \textcolor{black}{1400}    &  \textcolor{black}{1400}    & \textcolor{black}{1400}    & \textcolor{black}{1000}    \\ \hline
\end{tabular}}
\caption{Prices in euros of insurance on an ADS depending on portfolio.}
\label{tab:ins_portfolio}
\end{table}
 Since these products have no effect on access
 and non-protection probabilities, when referring to these probabilities, we do not 
 differentiate depending on the product considered, as they will remain the same regardless of the insurance adopted. 

\subsection*{A.2 Non-targeted attacks}

This section discusses the distribution and parameters used to model non-targeted attacks that may potentially threaten the ADS system. We model the arrival process, access and protection probabilities, and the impact of specific attacks, given the portfolio. 

\subsubsection*{A.2.1 Denial of Service attack features} 

Let us specify the distributions modeling the relevant parameters when the ADS is affected by a DoS attack.

\paragraph{Arrival process.} The number of potential DoS attacks within a one-year horizon is modeled through a Poisson distribution with mean \textcolor{black}{32}.\footnote{Based on the number of  organizations in the transportation sector  reported to have fallen victim to DoS attacks in the USA during 2022 \citep{crimereport}.}

\paragraph{Access probabilities.}  As stated, regardless of portfolio $\textbf{c}$, we assume the same distribution for ($p_{P}(\textbf{c})$,  $p_{L}(\textbf{c})$, $p_{P,L}(\textbf{c})$). These probabilities are sampled from a Dirichlet \textcolor{black}{$Dir(4,8,1)$} distribution, as the location system is more prone
to being attacked through a DoS attack than the perception system. 
Such type of attack is more likely to occur through the V2I system via an infected fixed element, say traffic light, than through a component of the perception system. We posit an attack occurring through both blocks as less likely, 
since it would entail a more complex and elaborate attack.

\paragraph{Non-protection probabilities.} We detail now the PNPs 
given the portfolios.

\begin{itemize}
\item \textit{None.} We assume that $q_{P}(0,0,0)\sim Beta(\textcolor{black}{27,3})$. Among other things, 
this entails that the expected probability of not protecting the perception system
when no additional measures are introduced is $\textcolor{black}{27/(27+3)}$.
Similarly, assume $q_{L}(0,0,0) \sim Beta(\textcolor{black}{26,3})$, $q_{D,P}(0,0,0)\sim Beta(\textcolor{black}{25,3})$ and $q_{D,L}(0,0,0)\sim Beta(\textcolor{black}{24},\textcolor{black}{3})$.  

\item \textit{$FwGw.$} When $FwGw$ is implemented,  PNPs are distributed as \textcolor{black}{$q_{P}(1,0,0)\sim Beta(5,65)$}, \textcolor{black}{$q_{L}(1,0,0) \sim Beta(4,65)$}, \textcolor{black}{$q_{D,P}(1,0,0)\sim Beta(3,65)$} and \textcolor{black}{$q_{D,L}(1,0,0)\sim Beta(2,65)$}. 
This entails e.g.\ that the expected probability of not protecting the perception
system when FwGW is implemented is $\textcolor{black}{5/(5+65)}$, thus importantly improving 
security against such threats.

\item \textit{$PmV.$} The  PNPs when a $PmV$ is used are assessed as  \textcolor{black}{$q_{P}(0,0,1)\sim Beta(5,95)$}, \textcolor{black}{$q_{L}(0,0,1)\sim Beta(4,95)$},  \textcolor{black}{$q_{D,P}(0,0,1)\sim Beta(3,95)$} and \textcolor{black}{$q_{D,L}(0,0,1)\sim Beta(2,95)$}.

\item \textit{FwGw, PmVs.} When both protection measures are implemented, the PNPs are modeled as  \textcolor{black}{$q_{P}(1,0,1)\sim Beta(5,125)$}, \textcolor{black}{$q_{L}(1,0,1)\sim Beta(4,125)$,  $q_{D,P}(1,0,1)\sim Beta(3,125)$ and $q_{D,L}(1,0,1)\sim Beta(2,125)$}.
\end{itemize}

\paragraph{Impacts.} The impacts of a DoS attack depend on the implemented 
defense. 

\begin{itemize}
    \item \textit{Financial.} All financial impacts are modeled using $Gamma$ distributions
    in Keuros. 
    \begin{itemize}
        \item \textit{None.} The impacts when no additional countermeasure 
        is introduced would be modeled as \textcolor{black}{$l_{FI}(0,0,0)\sim Gamma(7,3)$} Keuros. This implies, for example, that the expected output is \textcolor{black}{21 Keuros}.
        
        \item \textit{FwGw.} The impacts would follow  \textcolor{black}{$l_{FI}(1,0,0)\sim Gamma(6,3)$} Keuros.
         \item \textit{PmVs.} The impacts are modeled 
         as \textcolor{black}{$l_{FI}(0,0,1)\sim Gamma(4,3)$} Keuros.
         \item \textit{FwGw, PmVs.} If both defenses are implemented, impacts are modeled as \textcolor{black}{$l_{FI}(1,0,1)\sim Gamma(3,3)$} Keuros.
    \end{itemize}

\item \textit{Downtime.} The distributions that model the impact on each component are:
    
    \begin{itemize}
    \item \textit{None.} Assume \textcolor{black}{$l_{DT_P}(0,0,0)\sim Gamma(14,2)$},
      \textcolor{black}{$l_{DT_L}(0,0,0)\sim Gamma(15,2)$} and \textcolor{black}{$l_{DT_D}(0,0,0)\sim Gamma(18 ,3)$} hours. For instance, for the perception block, this entails that the expected downtime is \textcolor{black}{28 hours}.

    \item \textit{FwGw.} The distributions are \textcolor{black}{$l_{DT_P}(1,0,0)\sim Gamma(12,2)$},  \textcolor{black}{$l_{DT_L}(1,0,0)\sim \\Gamma(13,2)$} and \textcolor{black}{$l_{DT_D}(1,0,0)\sim Gamma(16,3)$} hours. 
    \item \textit{PmVs.} In this case, impacts are distributed as  \textcolor{black}{$l_{DT_P}(0,0,1)\sim Gamma(9,2)$}, \\ \textcolor{black}{$l_{DT_L}(0,0,1)\sim Gamma(10,2)$} and \textcolor{black}{$l_{DT_D}(0,0,1)\sim Gamma(13,3)$} hours. 
     \item \textit{FwGw,PmVs}.  The impacts would be distributed as \textcolor{black}{$l_{DT_P}(1,0,1)\sim Gamma(8,\\2)$},  \textcolor{black}{$l_{DT_L}(1,0,1)\sim Gamma(9,2)$} and \textcolor{black}{$l_{DT_D}(1,0,1)\sim Gamma(12,3)$} hours. 
    \end{itemize}
\end{itemize}


\subsubsection*{A.2.2 Software/hardware supply chain threat features.} 
\label{ref:section_sct}

This section presents the models associated with a software/hardware SCT attack.

\paragraph{Arrival process.} The annual number of potential attacks resulting from supply chain threats follows a Poisson distribution with  parameter  \textcolor{black}{2.75}.\footnote{Derived from reported number of accidents associated with models from a specific brand by the National Highway Traffic Administration (NHTSA) when employing an Advanced Driver Assistance System (ADAS) \url{https://static.nhtsa.gov/odi/inv/2021/INOA-PE21020-1893.PDF}, presuming that an attacker could exploit the same vulnerability in the vehicle software at a similar rate.}

\paragraph{Access probabilities.} Consider equally likely that a perpetrator will target either of the two entry points and less likely that it will attack both simultaneously. As a result, ($p_{P}(\textbf{c})$, $p_{L}(\textbf{c})$, $p_{P,L}(\textbf{c})) $  are modelled as \textcolor{black}{$ Dir(5,5,1)$}. It is assumed that the attacker faces the same difficulty level in accessing the system through a component regularly updated in the perception system, such as the V2V, or in the location block, such as the GPS. Otherwise, we assume it is less likely that it will be attempted to access the system through an update that affects components in both entry blocks, as it would require a much more elaborate attack.

\paragraph{Non-protection probabilities.} Table \ref{tab:acess_SCT} displays the parameters of the \emph{Beta} distributions employed to model  
PNPs given the implemented 
portfolio. 

\begin{table}[H]
\centering
\textcolor{black}{
\begin{tabular}{ccccc}
\hline
Portfolio & $q_{P}$ & $q_{L}$ & $q_{D,P}$ & $q_{D,L}$ \\ \hline
(0,0,0)   & (35,3)      & (33,3)      & (32,3)        & (31,3)        \\
(1,0,0)   & (5,10)      & (4,10)      & (3,10)        & (2,10)        \\
(0,1,0)   & (5,45)      & (4,45)      & (3,45)        & (2,45)        \\
(0,0,1)   & (5,30)      & (5,30)      & (3,30)        & (2,30)        \\
(1,1,0)   & (5,75)      & (4,75)      & (3,75)        & (2,75)        \\
(0,1,1)   & (5,105)      & (4,105)      & (3,105)        & (2,105)        \\
(1,0,1)   & (5,90)      & (4,90)      & (3,90)        & (2,90)        \\
(1,1,1)   & (5,125)      & (4,125)      & (3,125)        & (2,125)        \\ \hline
\end{tabular}}
\caption{SCT PNP distribution parameters depending on portfolio.}
\label{tab:acess_SCT}
\end{table}

\paragraph{Impacts.} Table \ref{tab:imp_SCT} quantifies the financial impacts ($l_{FI}$, $l_{ED}$,$l_D$) of an SCT  through the specified Gamma  distributions.

\begin{table}[H]
\centering
\textcolor{black}{
\begin{tabular}{ccccccccc}
\hline
\multirow{2}{*}{Portfolio} & \multirow{2}{*}{$l_{FI}$} & \multicolumn{3}{c}{$l_{ED}$} & \multicolumn{3}{c}{$l_{D}$} & \\ \cline{3-8}
                           &                           & $P$      & $L$     & $D$     & $P$     & $L$     & $D$                               \\ \hline
(0,0,0)                    & (10,2)                   & (9,2)  & (8,2) & (10,2) & (17,2) & (14,2) & (16,2)                 \\
(1,0,0)                    & (8,2)                   & (8,2)  & (7,2) & (9,2) & (12,2) & (11,2) & (13,2)      \\
(0,1,0)                    & (6,2)                   & (6,2)  & (5,2) & (7,2) & (10,2) & (9,2) & (11,2)                   \\
(0,0,1)                    & (7,2)                   & (7,2)  & (6,2) & (8,2) & (11,2) & (10,2) & (12,2)                    \\
(1,1,0)                    & (3,1)                   & (4,2)  & (3,2) & (5,2) & (4,2) & (5,2) & (6,2)                   \\
(0,1,1)                    & (2,1)                   & (3,2)  & (2,2) & (4,2) & (3,2) & (4,2) & (5,2)                   \\
(1,0,1)                    & (4,1)                   & (5,2)  & (4,2) & (6,2) & (5,2) & (6,2) & (7,2)                   \\
(1,1,1)                    & (1,1)                   & (2,1)  & (1,1) & (3,1) & (1,2) & (2,2) & (3,2)                   \\ \hline
\end{tabular}}
\caption{Impact distribution parameters of SCTs depending on portfolio.}
\label{tab:imp_SCT}
\end{table}

\subsection*{A.3 Targeted attacks}

This section delves into modeling specifics of targeted attacks, therefore
requiring the assessment of attackers' motivations to determine whether it is advantageous for the attackers to target the system under study, as 
 Section \ref{sec:targeted_att} explained. We thus provide details 
  to construct the attackers utility functions, taking into account the corresponding 
  uncertainties as well as the attack arrival processes and success probabilities. Subsequently, we discuss distributions to describe PNPs and 
  attack impacts.  The chosen distributions will reflect that the cyberterrorist group
  is better skilled than the criminal gang. 

\subsubsection*{ A.3.1 Attacker utility functions}
\label{sec:att_u_func}

This section outlines the construction of the attackers utility functions,
 first discussing their generic objectives, then the parametric form adopted and
 finally delving into the uncertainties about their preferences and behavior.

\paragraph{Attackers objectives.} The attackers may pursue the following objectives:

\begin{itemize}
\item \textit{Maximizing notoriety $nt$.} An attacker 
strives to gain influence to be later used in search of geopolitical objectives. 

\item \textit{Minimizing detection costs $c_d$.} These are the costs associated to the attacker being identified, including economic sanctions and/or legal condemnation, which could even lead to 
 their disappearance. 

\item \textit{Maximizing the sensitive information $s$ obtained.} This refers to the quantity of relevant data (valuable business information, customers' personal data,...) that an attacker illicitly obtains to sell for an economic gain. 

\end{itemize}

 \noindent The cyberterrorist group wants to maximize its notoriety. The criminal gang aims   to steal as much sensitive information as possible. Both attackers aim to minimize  detection costs. The costs of implementing wireless jamming or AML attacks are    negligible; thus, 
 attackers will not consider the implementation expenses involved in carrying out
 such cyberattacks.

\paragraph{Utility parametric form.} Assume the attackers preferences are modelled with the following piecewise risk-prone utility function

\begin{equation*}
u_A(y,a,\textbf{c}) =
    \text{exp}(h_A\times (y \times \pi - c_d)),
    \end{equation*}
\noindent with $y$ representing the success of the attack (1, if successful; 0, otherwise), 
$\pi$ denoting the variable that each attacker aims to maximize ($\pi_{Cy} = nt$, $\pi_{Cr} = s$),  and $h_A$ the risk proneness parameter. The defender's lack of knowledge about the attacker's preferences leads to the 
random utility model 
\begin{equation*}
U_A(y,a,\textbf{c}) =
    \text{exp}(H_A\times (y \times \Pi - C_d))
    \end{equation*}
where $\Pi$ and $C_d$ designate the random variables incorporating the uncertainties over the attacker's objectives. Assume \textcolor{black}{$H _A \sim  \mathcal{U}(1 \times 10^{-6}, 2 \times 10^{-6})$} models the 
uncertainty about the risk proneness (for both adversaries). 

\pagebreak
\paragraph{Specificities for cyberterrorist group}

\subparagraph{Notoriety modeling.}

As a proxy for notoriety, we aggregate the financial impact and number of deaths that the attacker is predicted to cause with an attack, that is, we use  $nt  = l_{FI}^A + l_{FA}^A \times VSL$, where  $l_{FI}^A$ represents the financial impacts that the attacker anticipates causing to the targeted ADS and $l_{FA}^{A}$ is the number of deaths expected by the attacker, and $VSL$ is \textcolor{black}{the value of statistical life (VSL), equivalent to 6 million euros}.\footnote{\textcolor{black}{Based on the VSL for Spain \citep{viscusi2020pricing} and an exchange rate euro-dollar of 0.9. VSL estimates changes in mortality risk in monetary terms. The term VSL is easily misinterpreted as the financial value of a person's life \citep{cameron2010vsl}; however, there is still no consensus on rewording \citep{freeman2014vsl}.
}}  We assume \textcolor{black}{$l_{FI}^{A} \sim \text{Gamma} (\omega \times 8, \omega \times 3)$} and \textcolor{black}{$l_{FA}^{A} \sim \text{Poisson}(\omega \times 25)$}. Table \ref{tab:b_param} displays the value of $\omega $ depending on whether the cyberterrorist group makes a wireless jamming or AML attack and the 
  implemented portfolio. For instance, when no defenses are implemented  ({\bf c}=(0,0,0)), the expected financial impacts for wireless jamming are \textcolor{black}{$0.8 \times 8 \times 3 =  19200$} euros, and the expected number of deaths is \textcolor{black}{$0.8 \times 25 =  20$}.
\begin{table}[h]
\centering
\textcolor{black}{
\begin{tabular}{cccc}
\hline
 Portfolio & $AML\_at$ & $wir\_jam$  \\ \hline
(0,0,0)   & 1        & 0.8        \\
(1,0,0)   & 1        & 0.6          \\
(0,1,0)   & 0.2        & 0.8          \\
(0,0,1)   & 1        & 0.4        \\
(1,1,0)   & 0.2        & 0.6       \\
(0,1,1)   & 0.2        & 0.4      \\
(1,0,1)   & 1        & 0.2        \\
(1,1,1)   & 0.2        & 0.2      \\ \hline
\end{tabular}}
\caption{Values of parameter $\omega$ for different portfolio configurations and attacks.}
\label{tab:b_param}
\end{table}
Acknowledging the lack of information about the other companies, we assume that \textcolor{black}{$l_{FI}^{A} \sim \text{Gamma} (7.2 , 2.7)$} and \textcolor{black}{$l_{FA}^{A} \sim \text{Poisson}(22.5)$} for an AML attack when targeting the other companies. Similarly, in the case of a wireless jamming attack to other companies, we assume \textcolor{black}{$l_{FI}^{A} \sim \text{Gamma} (5.6, 2.1)$} and \textcolor{black}{$l_{FA}^{A} \sim \text{Poisson}(17.5)$}.

\subparagraph{\textit{Detection costs and probability.}} Denote by $p^{det}_{(1,j,k)}$
the detection probability of attack $j$ targeting block $k$ of our ADS, and 
$p^{det}_{(i,j)}$, the detection probability of attack $j$ targeting company $i$'s ADS, \textcolor{black}{$i=2,3$}.   We assume for the AML attack that \textcolor{black}{$p^{det}_{(1,AML\_at,k)}\sim Be(2,998), k\in \{L,P\}$}  and \textcolor{black}{$p^{det}_{(1,AML\_at,(L,P ))}\sim Be(1,99)$}. Regarding the wireless jamming attack, \textcolor{black}{$p^{det}_{(1,wir\_jam,k)}\sim  Be(1,999),k\in \{L,P\}$}  and \textcolor{black}{$p^{det}_{(wir\_jam,(L,P))}\sim Be(5,995)$}. Acknowledging the lack of information about the other companies, we assume that \textcolor{black}{$p_{(i,j)}^{det}\sim Be(6,994),$} $i\in \{2,3\}$, $j\in \{AML\_at, wir\_jam\}$.  If either of the two attacks is detected, $cost_{det}$ will be modeled as \textcolor{black}{$\mathcal{U}(100000, 130000)$}. We then compute the expected detection costs $c_d$ as $p^{det}_{(1,j,k)} \times {cost}_{det}$ or $p^{det}_{(i,j)} \times {cost}_{det}$ depending on the targeted system.

\paragraph{Specificities for criminal gang}

\subparagraph{\textit{Sensitive information modeling.}}
Let $v_s$ be the value of a data record containing sensitive information and $N_s$, the count of stolen data records. The gain from selling stolen sensitive information will be $N_s \times v_s$. To account for uncertainty, the value of a data record is  
modelled as \textcolor{black}{$v_s \sim \mathcal{U}(0.8\times 100, 1.2\times 
  100)$}. To estimate the number of data records stolen in a successful attack, consider that \textcolor{black}{$N_s\sim \mathcal{U}(0,ft_{records})$}, being \textcolor{black}{$ft_{records} = 660$} with the default portfolio and \textcolor{black}{$ft_{records} = 360$} when the AML is implemented. 
  Recognizing the absence of information on the other companies, we assume that \textcolor{black}{$ft_{records}$ = 600} when the other companies' vehicles are attacked.

\subparagraph{\textit{Detection costs and probability.}} It is considered that \textcolor{black}{$p^{det}_{(1,AML\_{at},k)}\sim Be(3,997),\\ k\in \{L,P\}$}, \textcolor{black}{$p^{det}_{(1,AML\_{at},(L,P))}\sim Be(5,995)$} and \textcolor{black}{$p^{det}_{(i,AML\_at)}\sim Be(4,996)$}.  Detection costs are modeled as  \textcolor{black}{${costs_{det}}\sim \mathcal{U}(100000,130000)$}.  As with the cyberterrorist group, these values are employed to calculate the expected costs $c_d$ as  \textcolor{black}{$p^{det}_{(AML\_{at},k)} \times costs_{det}$} or \textcolor{black}{$p^{det}_{(i,AML\_at)} \times costs_{det}$, $i=2,3$}. 

\subsubsection*{A.3.2 Attack arrival process and success probability}
\paragraph{Arrival process.} We model the attack arrival process from 
 potential attackers through Poisson processes with parameters 
 determined using expert knowledge. In particular, for the cyberterrorist group, the number of potential attacks within a one-year period is modelled using Poisson process with rate \textcolor{black}{4}; for the criminal gang, the Poisson rate is
  \textcolor{black}{3}. 

\paragraph{Successful attack probabilities.} We assess the attackers' beliefs that an
attack is successful given a certain portfolio. When the cyberterrorist group conducts an AML attack:

\begin{itemize}

\item When no defense is applied \textcolor{black}{$P_{A}(Y =1|a^1_{(AML\_at,P)}, (0,0,0)) \sim Be(75,25)$,} 
being \textcolor{black}{$P_{A}(Y =1|a^1_{(AML\_at,P)}, (0,0,0))$} the probability that the attack is successful when the attacker launches an AML attack to block $P$ of the system, given that the portfolio (0,0,0) has been implemented. Similarly, \textcolor{black}{$P_{A}(Y =1|a^1_{(AML\_at,L)},  (0,0,0))  \sim Be(75,25),\\ P_{A}(Y =1|a^1_{(AML\_at,(P,L))}, (0,0,0))\sim Be(90,10)$}.  We assume the probability of successful attack is higher when both entries are attacked, despite of being more easily detected.

\item When the AML protection is implemented \textcolor{black}{$P_{A}(Y =1 |a^1_{(AML\_at,P)}, (0,1,0))$, \\$P_{A}(Y =1 |a^1_{(AML\_at,L)}, (0,1,0))\sim Be(15,85)$,$P_{A}(Y =1 |a^1_{(AML\_at,(P,L))}, (0,1,0))\sim \\  Be(10,90)$.}
\end{itemize}

\noindent As the previous distributions show, implementing a defensive measure not only reduces the PNPs, but would also act as a deterrent for the attacker, since the attacker’s beliefs over a successful attack decreases. Security curves, Figure \ref{fig:sub1a}, can be used to determine the distribution parameters. In turn, Table \ref{tab:wir_jam_succ_at_cyg} displays the parameterization when the attacker carries out a  wireless jamming learning attack.

\begin{table}[H]
\centering
\textcolor{black}{
\begin{tabular}{cccc}
\hline
Portfolio & \textit{P} & \textit{L} & \textit{P,L} \\ \hline
(0,0,0)   & (60,40)    & (60,40)    & (70,30)      \\
(1,0,0)   & (25,75)    & (25,75)    & (35,65)      \\
(0,0,1)   & (25,75)    & (25,75)    & (35,65)      \\
(1,0,1)   & (5,95)     & (5,95)     & (15,85)      \\ \hline
\end{tabular}}
\caption{\textcolor{black}{Successful attack probability beta parameters of wireless jamming conducted by the cyberterrorist group regarding the targeted block.}}
\label{tab:wir_jam_succ_at_cyg}
\end{table}

\noindent \textcolor{black}{Recognizing a lack of information, we model the attack probability of the cyberterrorist group on a company $i$'s vehicle as $\textcolor{black}{P_{A}(Y = 1|a^{i}_j)\sim \text{Be}(1,1), \textcolor{black}{i\in\{ 2,3\}}}$, \textcolor{black}{$j\in$} $\textcolor{black}{\{AML\_at,wir\_jam \}}$, under the current status.}

\textcolor{black}{Similarly, Table \ref{tab:suc_beta_aml_crg} shows the parameters of the beta distributions modeling the probabilities of a successful AML attack carried out by the criminal gang, given the portfolio implemented regarding the targeted block. As before, we model the probabilities of a successful AML attack on vehicles from other companies as $P_{A}(Y = 1|\textcolor{black}{a_{AML\_at}^{i})\sim \text{Be}(1,1), i\ = 2,3}$.}

\begin{table}[H]
\centering
\textcolor{black}{
\begin{tabular}{cccc}
\hline
Portfolio & \textit{P} & \textit{L} & \textit{P,L} \\ \hline
(0,0,0)   & (65,35)    & (65,35)    & (80,20)      \\
(0,1,0)   & (15,85)    & (15,85)    & (25,75)      \\ \hline
\end{tabular}}
\caption{\textcolor{black}{Successful attack probability beta parameters of AML attack conducted by criminal gang regarding the targeted block.}}
\label{tab:suc_beta_aml_crg}
\end{table}

\vspace{-1cm}
\textcolor{black}{\subsubsection*{A.3.3 Computation of attack probabilities}}

\noindent \textcolor{black}{To estimate the attack probabilities for each 
attacker, we conducted \textcolor{black}{10000} simulations using the previously outlined modeling details to derive vectors $\tau(\textbf{c})$ and $\Gamma_{j}^1(\textbf{c})$, as explained in Section \ref{sec:targeted_att}.
In particular, for the cyberterrorist group case, we assess the probability of implementing one of the following actions:}

\textcolor{black}{
\begin{itemize}
\item[-] [1-3] AML attack targeting the location block, perception block, or both $(a^1_{AML\_at,k})$.
\item[-] [4-7]Wireless jamming targeting the location block, perception block or both $(a^1_{wir\_jam,k})$.
\item[-] \textcolor{black}{[8-9] AML attack to  vehicles of other company $(a^i_{AML\_at},i=2,3)$}.
\item[-] \textcolor{black}{[9-10] Wireless jamming targeting vehicles of other company $(a_{wir\_jam}^i,i=2,3)$}.
\end{itemize}}

\noindent A similar analysis for the criminal gang was performed.

\subsubsection*{A.3.4 Non-protection probabilities}

This section describes the modeling details for parameters related to PNPs when the system is subject to a targeted attack. 
Note that the selection of PNPs for AML defenses can be accomplished through an analysis using security curves similar to those depicted in Figure \ref{fig:sub1a}.

\paragraph{Cyberterrorist group.}{Table \ref{erry} shows the \emph{Beta} distribution parameters modeling the PNP for 
each defense when an AML attack is perpetrated.}
\noindent \textcolor{black}{For instance, the parameters for $q_p$ associated with the portfolio $(0, 0, 0)$ indicate that the expected PNP is \textcolor{black}{72/(72+3)} for an AML attack conducted by the cyberterrorist group targeting the perception block.}

\begin{table}[!htbp]
\centering
\textcolor{black}{
\begin{tabular}{ccccc}
\hline
Portfolio & $q_{P}$  & $q_{L}$  & $q_{D,P}$ & $q_{D,L}$ \\ \hline
(0,0,0)   & (72,3)   & (73,3) & (74,3)    & (75,3)  \\
(0,1,0)   & (5,195) & (4,195)  & (3,195)  & (2,195)   \\ \hline
\end{tabular}}
\caption{Cy AML PNP distribution beta parameters depending on portfolio.}
\label{erry}
\end{table}

Analogously, Table \ref{ref:wir_jamm_q} displays the corresponding parameters in the event of an attack through wireless jamming. 

\begin{table}[!htbp]
\centering
\textcolor{black}{
\begin{tabular}{ccccc}
\hline
Portfolio & $q_{P}$ & $q_{L}$ & $q_{D,P}$ & $q_{D,L}$ \\ \hline
(0,0,0)   & (8,3)   & (9,4) & (10,3)  & (11,3)  \\
(1,0,0)   & (5,60)   & (4,60) & (3,60)   & (3,60)     \\
(0,0,1)   & (5,95) & (4,95) & (4,90)   & (3,90)   \\
(1,0,1)   & (5,115) & (4,115) & (3,115)   & (2,115)   \\ \hline
\end{tabular}}
\caption{Cy wjam PNP distribution beta parameters.} 
\label{ref:wir_jamm_q}
\end{table}

\paragraph{Criminal gang.} Table \ref{tab:cg_aml} shows the \emph{Beta} distribution parameters modeling the PNP for 
each defense against an AML attack.
\begin{table}[h]
\centering
\textcolor{black}{
\begin{tabular}{ccccc}
\hline
Portfolio & $q_{P}$  & $q_{L}$ & $q_{D,P}$ & $q_{D,L}$ \\ \hline
(0,0,0)   & (16,5)   & (15,5)  & (22,5)    & (20,5)    \\
(0,1,0)   & (5,245) & (4,245) & (3,245)  & (2,245)   \\ \hline
\end{tabular}}
\caption{Cr AML portfolio dependent PNP beta parameters.}
\label{tab:cg_aml}
\end{table}

\subsubsection*{A.3.5 Attack impacts}

Concerning the impact of targeted attacks,
  we discern when the system is targeted between attacks originating 
  from a cyberterrorist organization or a criminal gang, as cyberterrorists are assumed to be
more skilled.  

\paragraph{Cyberterrorist group.} Table \ref{tab:aml_imp} displays the parameters of the Gamma and Poisson distributions modelling
 the impacts of adversarial attacks perpetrated by a cyberterrorist group, both with and without AML protection in place.

\begin{table}[H]
\centering
\textcolor{black}{
\begin{tabular}{cccccccc}
\hline
\multirow{2}{*}{Portfolio} & \multirow{2}{*}{$l_{FI}$} & \multicolumn{3}{c}{$l_{ED}$} & \multicolumn{3}{c}{$l_{D}$} \\ \cline{3-8} 
                           &                           & $P$      & $L$     & $D$     & $P$     & $L$     & $D$     \\ \hline
(0,0,0)                    & (12,2)                    & (6,2)    & (7,2)   & (6,3)   & (30,2)  & (36,2)  & (45,2)  \\
(0,1,0)                    & (2,2)                     & (1,2)    & (1,2)   & (1,3)   & (2,2)   & (3,1)   & (4,2)   \\ \hline
\end{tabular}}
\caption{Portfolio dependent impact distribution \textcolor{black}{gamma} parameters of AML attack by cyberterrorist.}
\label{tab:aml_imp}
\end{table}

\noindent Similarly, Table \ref{tab:impact_dist_wj} shows the parameters of the distributions modelling the impacts caused by a wireless jamming attack.

\begin{table}[H]
\centering
\textcolor{black}{
\begin{tabular}{cccccccc}
\hline
\multirow{2}{*}{Portfolio} & \multirow{2}{*}{$l_{FI}$} & \multicolumn{3}{c}{$l_{ED}$} & \multicolumn{3}{c}{$l_{D}$} \\ \cline{3-8} 
                           &                           & $P$      & $L$     & $D$     & $P$     & $L$     & $D$     \\ \hline
(0,0,0)                    & (13,2)                    & (4,2)    & (4,2)   & (5,2)   & (22,2)  & (21,2)  & (23,2)  \\
(1,0,0)                    & (6,2)                     & (3,2)    & (3,2)   & (4,2)   & (16,2)  & (15,2)  & (17,2)  \\
(0,0,1)                    & (3,2)                     & (2,2)    & (2,2)   & (2,2)   & (9,2)   & (8,2)   & (10,2)  \\
(1,0,1)                    & (1,1)                     & (1,2)    & (1,2)   & (1,2)   & (2,2)   & (2,2)   & (3,2)   \\ \hline
\end{tabular}}
\caption{Portfolio dependent impact distribution \textcolor{black}{gamma} parameters of wireless jamming conducted by cyberterrorist group.}
\label{tab:impact_dist_wj}
\end{table}

\paragraph{Criminal gang.} Table \ref{tab:impact_dist_cg_aml} displays the distribution parameters of the impacts of an  AML attack carried out 
by a criminal gang.

\begin{table}[H]
\centering
\textcolor{black}{
\begin{tabular}{ccccc}
\hline
\multirow{2}{*}{Portfolio} & \multirow{2}{*}{$l_{FI}$} & \multicolumn{3}{c}{$l_{ED}$} \\ \cline{3-5} 
                           &                           & $P$      & $L$     & $D$     \\ \hline
(0,0,0)                    & (10,2)                    & (5,2)    & (5,2)   & (4,2)   \\
(0,1,0)                    & (2,2)                     & (1,2)    & (1,2)   & (1,3)   \\ \hline
\end{tabular}}
\caption{Portfolio dependent impact distribution \textcolor{black}{gamma} parameters of AML attack by a criminal gang.}
\label{tab:impact_dist_cg_aml}
\end{table}

\end{document}